\documentclass[10pt]{article} 
\usepackage[preprint]{neurips_2026/neurips_2026}

\usepackage{wrapfig}
\usepackage{listings}
\usepackage[disable]{todonotes}
\usepackage{tcolorbox}
\usepackage{subcaption}
\usepackage{float}
\usepackage{amsmath, amsthm, amssymb}
\usepackage[T1]{fontenc}
\usepackage[utf8]{inputenc}
\usepackage{hyperref}
\usepackage{enumitem}
\usepackage{tikz}
\usepackage[table,dvipsnames]{xcolor}
\usetikzlibrary{shapes.geometric, positioning, decorations.pathmorphing}
\usepackage{float}
\usepackage{listings}
\usepackage{dsfont}
\usepackage{mathtools}
\usepackage{booktabs}
\colorlet{LightLimeGreen}{LimeGreen!30}

\newtheorem{prop}{Proposition}

\newcommand{\PUJ}[1]{\textcolor{green}{{\bf PUJ:} #1}}

\hypersetup{hidelinks}
\hypersetup{
  colorlinks   = true, 
  urlcolor     = blue, 
  linkcolor    = blue, 
  citecolor   = SeaGreen 
}

\title{Rethinking Cross-Lingual Gaps\\ via Response Variance}

\author{
      Vihari Piratla\\ \texttt{vpiratla@google.com} \\
      Google DeepMind 
      \And
      Purvam Jain \\
      Google DeepMind
      \And
      Darshan Singh  \\
      Google DeepMind 
      \AND
      Trevor Cohn \\
      Google Research
      \And
      Preethi Jyothi  \\
      Google DeepMind
      \And
      Partha Talukdar \\
      Google DeepMind}

\newcommand{\mmlu}{MMLU (with mixup)}
\newcommand{\eclektic}{ECLeKTic}
\newcommand{\neclektic}{Year-ECLeKTic}

\begin{document}

\maketitle


\begin{abstract}
Any piece of knowledge is usually expressed in one or a handful of natural languages on the web or in any large corpus. Large Language Models (LLMs) act as a bridge by acquiring knowledge from a \emph{source} language and making it accessible when queried using \emph{target} languages. A \emph{cross-lingual gap} is a drop in accuracy incurred when querying knowledge in a target language rather than the source language.
Existing research focused on modeling or training failures leading to cross-lingual gaps.
In this work, we take an alternative view to characterize the nature of cross-lingual error, and hypothesize that the variance of responses in the target language is a key cause of this gap. For the first time, we formalize the cross-lingual gap in terms of biased and unbiased errors.
We empirically validate our hypothesis through multiple inference-time interventions that control variance and reduce the cross-lingual gap. We demonstrate a few test-time ensemble methods that reduce response variance, and thereby improve source-target transfer scores by up to 12 absolute points yielding relative gains of 8\% to over 50\% across various LLMs.      

\end{abstract}

\section{Introduction}
\label{sec:intro}
Large Language Models (LLMs) have revolutionized information access. Central to LLM's mission is to assimilate knowledge universally and make it available at large without any language barriers. State-of-the-art LLMs are multilingual: Gemini 
supports over 40 languages~\citep{Geminisupport}, GPT-5 supports at least 12 languages~\citep{GPT5-system-card} (with no official number of supported languages) and open-source models like Gemma-3 support over 100 spoken languages~\citep{Gemma3card}.
Since pretraining data cannot contain duplicate information for every language, cross-lingual generalization is a necessary capability for LLMs. However, LLMs are known to have a disparity in recalling knowledge across languages~\citep{jiang2020x,kassner2021multilingual,qi2023cross,mmlu_mixup,goldman2025eclektic}.

Our main objective is to understand the causes of poor transfer of parametric knowledge across languages. Therefore, we evaluate models on knowledge-intensive tasks in a closed-book QA setting, i.e., without access to tools such as grounding in search. Cross-lingual gaps are quantified by assessing performance on parallel datasets with varying language-specific surface forms of the prompts. We view this parallel data through two evaluation perspectives, \emph{source} and \emph{target} languages. Prompts in the source split are (roughly) in-distribution with respect to pretraining data, while those in the target split are considered out-of-distribution. Parity between source and target is achieved when the model generalizes across languages.
For instance, consider a question derived from a Wikipedia article that is only available in Hindi: {\it When was Kreeda Bharti established?} Gemini-2.5-Flash (with thinking) correctly answers the question when posed in Hindi (source language) but the same question in Hebrew (a target language) is often answered incorrectly. 

\begin{figure}[t]
\begin{subfigure}[t]{0.36\linewidth}
  \includegraphics[width=\linewidth]{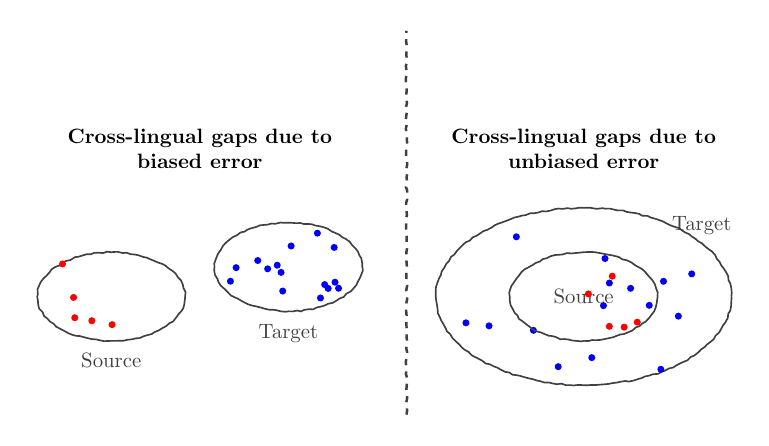}
\caption{Illustration of response distribution when the cross-lingual gaps are due to biased or unbiased error.}  
\label{illus_fig1a}
\end{subfigure}\hfill
\begin{subfigure}[t]{0.3\linewidth}
    \centering
    \includegraphics[width=\linewidth]{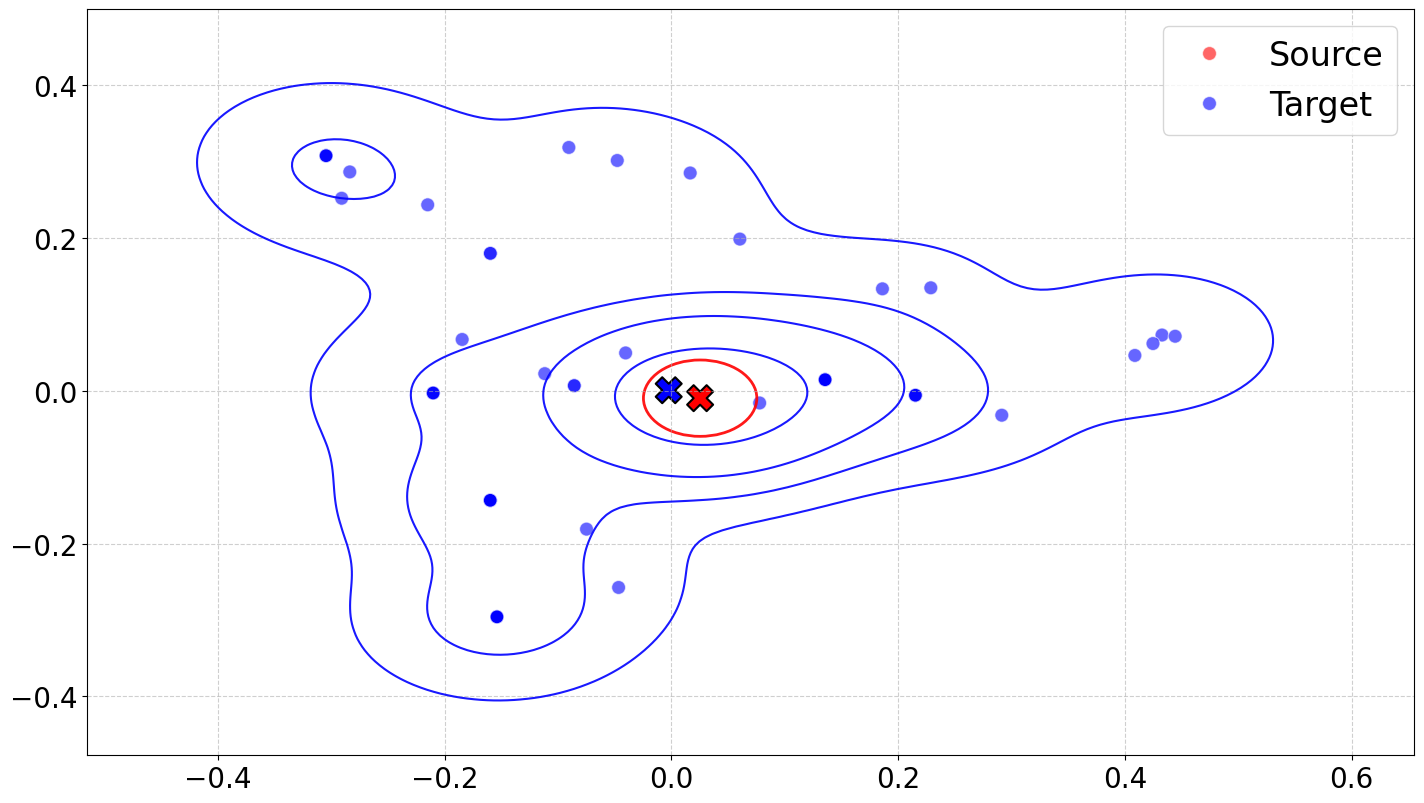}
    \caption{PCA projection of responses (G-2.5-Flash) in source and target for the English-sourced question: {\it What was the name of the protagonist in the Prizzi novels?}}
\end{subfigure}\hfill
\begin{subfigure}[t]{0.3\linewidth}
  \includegraphics[width=0.9\linewidth]{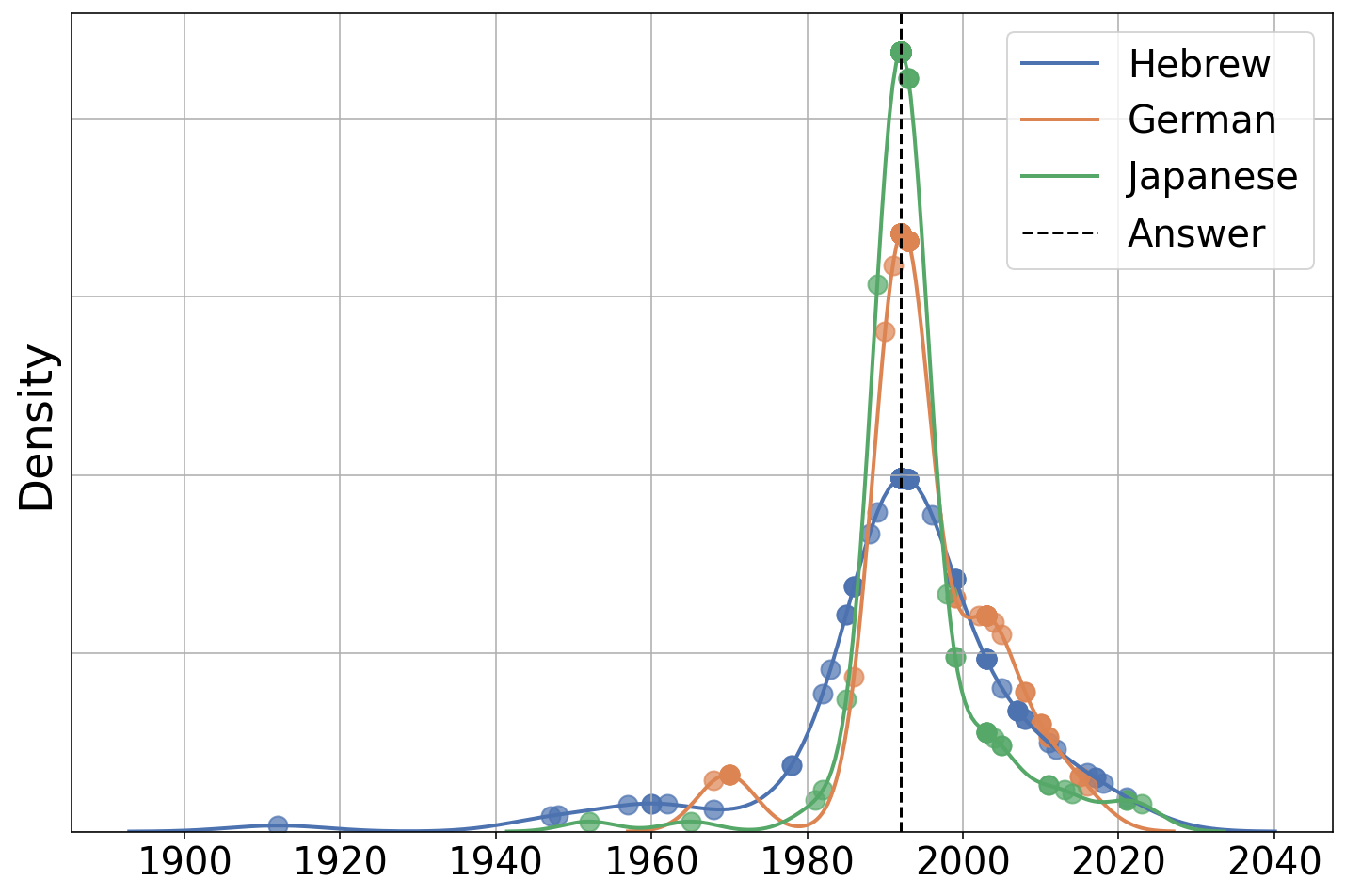}
  \caption{Distribution of hundred responses (with G-2.5-Flash) for the Hindi-sourced question ``When was Kreeda Bharti established?'' in various languages. 
  }    
\end{subfigure}
\caption{If the model has knowledge barriers, we expect target responses to be biased (left of (a)). In practice, we observe the target responses are distributed around source such that their respective average values coincide (right of (a)). In (b), we show PCA projections of source and target responses where each response is a dot, and crosses represent centroids. We show kernel density estimate fits for three target languages for a numerical question in (c).} 
\label{fig:problem_illustrated}
\end{figure}

Past work postulated many potential causes of cross-lingual gaps despite mounting evidence for language-agnostic representations~\citep{dumas2024llamas,schafer2024role,brinkmann2025large} in LLMs. 
We adopt a different approach and characterize cross-lingual error. In an extreme scenario when the parametric knowledge of LLMs is language-siloed, or when the model fails to comprehend a prompt in a foreign language, we expect the responses in the target language to be biased. On the other hand, if the model is simply under-confident in the target language, we expect errors to be zero-centered (unbiased) but accompanied by higher response variance in that language.  
We contextualize biased and unbiased cross-lingual error with an example question sourced from a Hindi document: {\it When was Kreeda Bharti established?} with the correct answer {\it 1992}.
If there is a knowledge barrier or the entity {\it Kreeda Bharti} is unrecognized in any language other than Hindi, we expect the model to respond with a random guess anywhere from 1500 BC to 2024 AD leading to significant {\it bias} relative to the Hindi response (1992). On the other hand, if the gap is due to variance alone, we expect target responses to be distributed more widely around the source response,  say 1992$\pm$30. Figure~\ref{illus_fig1a} sketches source and target responses for a hypothetical example when the cross-lingual gaps are due to biased (left) or unbiased (right) errors.

Prior work largely attributed cross-lingual gaps to knowledge barriers inducing biases in the target responses~\citep{mmlu_mixup,wang2024cl2cm}. The problem is more severe if the gaps are indeed due to biases because it requires rethinking LLM pretraining, tokenization, embeddings, etc. Crucially, the literature has often overlooked variance as an explanation for  cross-lingual gaps, which we address in this work. 
In contrast to the prevailing wisdom of attributing gaps to biases, we demonstrate the absence of bias in the two examples of Figure~\ref{fig:problem_illustrated}(b, c).%
\footnote{We present many examples with free-form or numeric responses, similar to Figure~\ref{fig:problem_illustrated}(b, c), in Appendix, Fig~\ref{fig:main_b_extended},~\ref{fig:main_c_extended}\todo[inline]{Purvam, please fill}.}
In Figure~\ref{fig:problem_illustrated}(b), we show PCA-projected embeddings of responses for a question sourced from an English document: {\it What was the name of the protagonist in the Prizzi novels?} with answer: {\it Charlie Partanna}. (More details about the embeddings are in Appendix~\ref{sec:embed_details}.) In Figure~\ref{fig:problem_illustrated}(c), we plot various responses to our running example: {\it When was Kreeda Bharti established?} in three languages. The question is answered correctly in Hindi with high confidence; we omit it from the plot to emphasize variance in the other languages. 
We make two observations based on Fig~\ref{fig:problem_illustrated}: (1) target responses exhibit higher variance, and (2) their mean coincides with the source response despite this high variance.


\begin{figure}[tb]
    \centering
 \begin{subfigure}[t]{.48\linewidth}
    \centering \includegraphics[width=\linewidth]{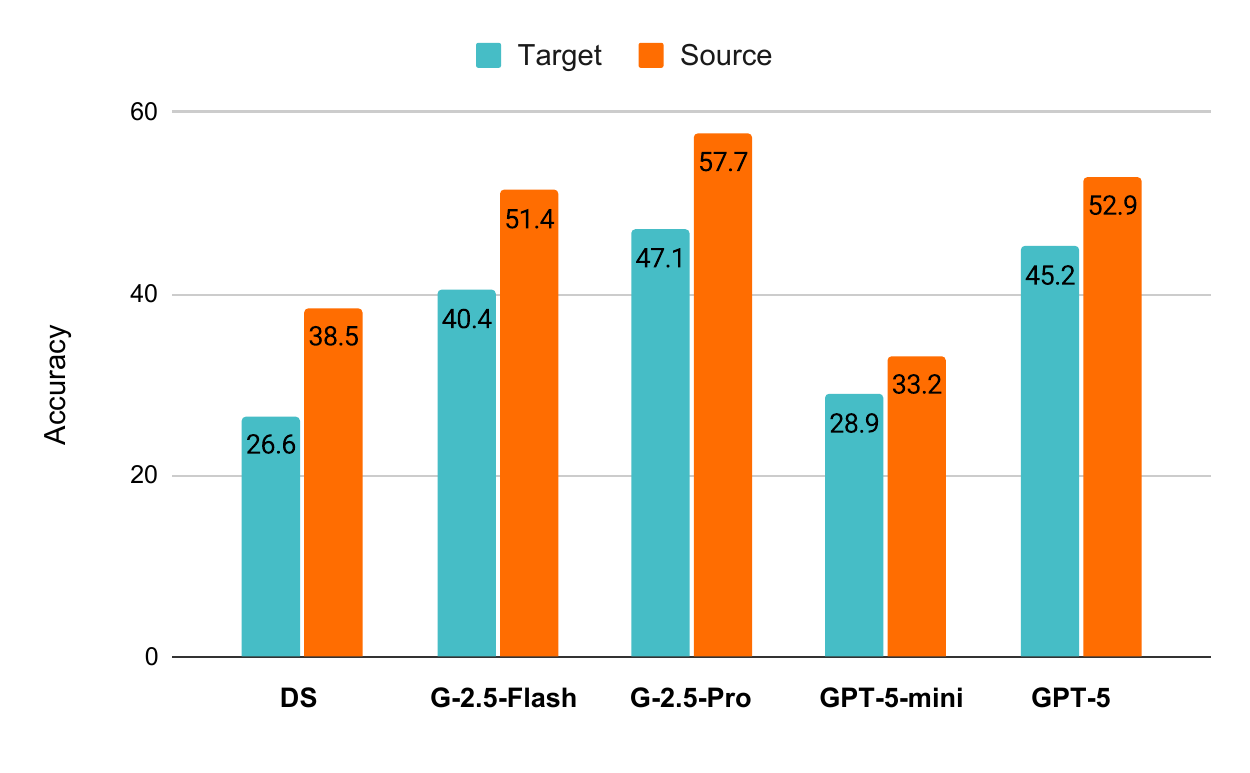}
    \caption{\eclektic{}}
  \end{subfigure}
  \begin{subfigure}[t]{.48\linewidth}
    \includegraphics[width=\linewidth]{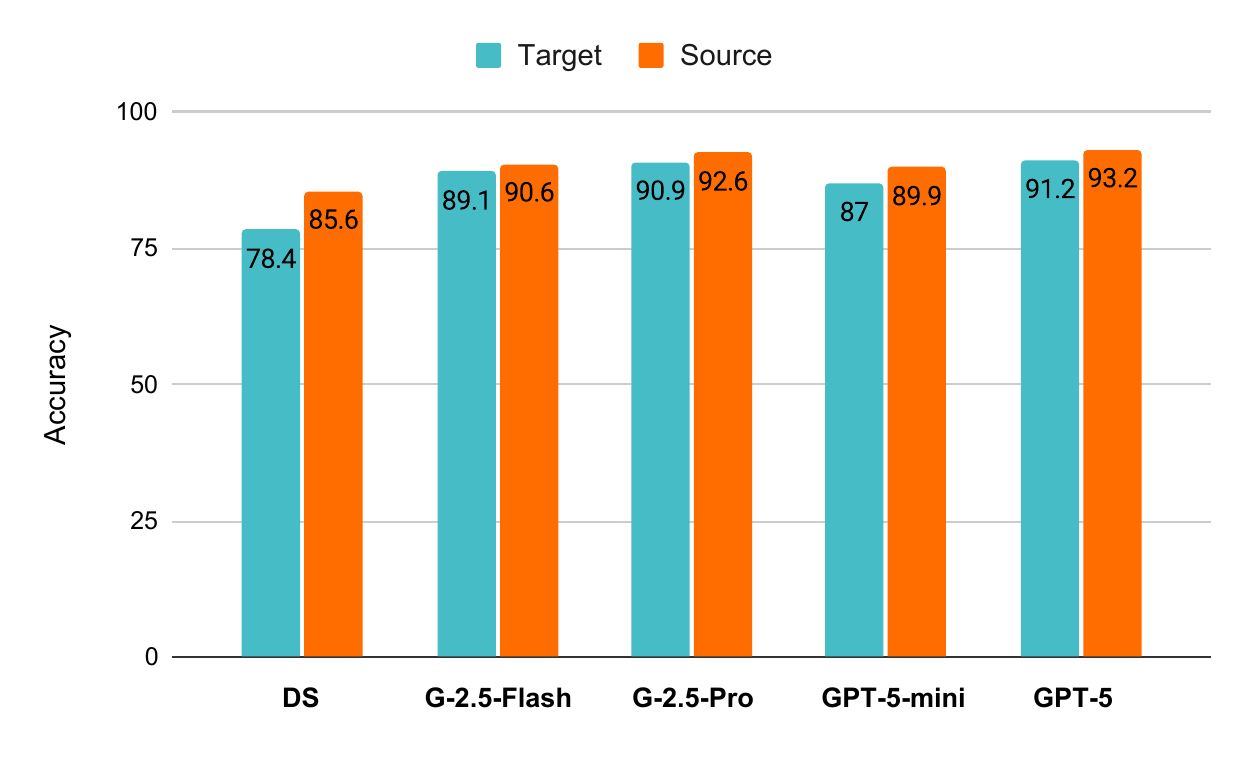}
    \caption{\mmlu{}}
  \end{subfigure}
    \caption{Cross-lingual performance gaps on \eclektic{} and \mmlu{} using different LLMs. Here, DS refers to the DeepSeek model. G-2.5-Flash, G-2.5-Pro are Flash and Pro flagship Gemini models and GPT-5-mini \& GPT-5 are Open AI models. All models exhibit significant cross-lingual gaps on the \eclektic{} benchmark.}
    \label{fig:xlang_gaps}
\end{figure}

Cross-lingual gaps persist across state-of-the-art LLMs, as shown in Figure~\ref{fig:xlang_gaps} on two recent benchmarks. Characterising the cause of such cross-lingual gaps is important to guide appropriate mitigation strategies.

\paragraph{Contributions:}
\begin{itemize}[itemsep=0pt, topsep=0pt, leftmargin=*]
    \item We study the nature of cross-lingual errors and validate for the first time that the gaps are dominantly due to unbiased error, i.e., confidence transferring poorly from source to target languages ($\S$\ref{sec:model}, $\S$\ref{sec:ensemble_expts}).
    \item We additionally demonstrate that (response) variance in source and target languages are proportional. As a consequence, source-target gaps diminish with decreasing variance in source ($\S$\ref{sec:confidence_vs_gaps}). 
    \item We empirically validate our claims across two benchmarks and five closed/open SoTA LLMs. We present some inference-time interventions that help mitigate cross-lingual gaps ($\S$\ref{sec:input_ensemble}).
\end{itemize}

\section{A Framework of Cross-lingual Gaps}
\label{sec:model}
In this section, we define how the target response distribution transforms under biased or unbiased cross-lingual errors. A major hurdle is in capturing the variance in responses. For a given model $M$ and prompt $x$, what is the distribution of responses $M(x)$? We cannot simply use logits to characterize the response distribution because (a) flagship models do not provide logits, (b) tokenization differences across languages complicates cross-lingual response probability comparisons. Besides, logits can vary between different uses of the same prompt with modern LLMs for various reasons including hardware issues such as floating point errors and/or expert routing errors.  

We work around the issue of capturing response variance by assuming a latent effective logit vector from which responses are drawn. We denote the effective logit vector by the unit vector $\vec{z}$ and its scale parameter, $\alpha$. The sampling process of a model $M$ when prompted with input $x$ emitting a response $\hat{y}$ is summarized below. 
\begin{align*}
    &\alpha\vec{z} \triangleq M(x), \quad \text{s.t. } \|\vec{z}\|=1,\alpha>0\\
    &\hat{y}\sim \operatorname{Categorical}(\operatorname{softmax}(\alpha\vec{z})).
\end{align*}

For the sake of analysis (in this section), we assume that the response space is enumerable and shared between source and target languages. We may achieve this by collating many responses in both source and target, and normalizing the unique values to only encode the underlying concept while ignoring the language. For instance, we normalize \{{\it order of santiago, order de santiago, ordem de santiago}\} to {{\it order of santiago}. Hereafter, we will treat the response space as categorical, with only the levels defined by the unique normalized values and logits as their corresponding scores.

With the response space normalized, we can quantify the probability of shared responses between source and target in the presence of response uncertainties. However, we must first model how the logit vector shifts from source to target. We denote the logit vector for source as $\vec{z}_s$ with scaling parameter $\alpha_s$. The target logit vector is expected to be unrelated to source if the cross-lingual errors are biased, i.e., the target logit vector for the same question, $\vec{z}_b$, is such that their modes do not match, i.e., $\arg\max \vec{z}_s\neq \arg\max\vec{z}_b$. If there is no bias but high variance, the target responses are expected to be distributed with a flatter logit distribution ($\alpha_t\vec{z}_s$) for some values of $\alpha_t> 0$. Since we do not know the relative contribution of bias and variance to cross-lingual gap, we model target responses as a mixture of both the distributions with an unknown mixing coefficient: $\pi, 0\leq\pi\leq 1$. Overall, our model of source and target responses is summarized below. 
\begin{figure}[H]
\begin{subfigure}[t]{0.35\textwidth}
\begin{align*}
&\text{\color{red}{Source response distribution:}} \\
&\alpha_s\vec{z}_s = M(x_s)\quad \text{s.t. } \|\vec{z}_s\|=1,\\
&\hat{y}_s \sim \operatorname{Categorical}\!\big(\operatorname{softmax}(\alpha_s\vec{z}_s)\big)
\end{align*}
\end{subfigure}
\begin{subfigure}[t]{0.62\textwidth}
\begin{align*}
&\text{\color{blue}{Target response distribution:}} \\
&\kappa \sim \operatorname{Bernoulli}(\pi), \\
&\vec{z}_t \mid \kappa
=
\begin{cases}
\alpha_t\vec{z}_s, & \kappa = 1 \quad \text{(unbiased error)} \\
\alpha_t\vec{z}_b, & \kappa = 0 \quad \text{(biased error)} \\
& \text{where } \arg\max \vec{z}_s\neq \arg\max \vec{z}_b, \|\vec{z}_b\|=1
\end{cases} \\
&\hat{y}_t \sim \operatorname{Categorical}\!\big(\operatorname{softmax}(\vec{z}_t)\big).
\end{align*}
\end{subfigure}
\end{figure}

Figure~\ref{fig:problem_illustrated} (a) illustrates the two scenarios of cross-lingual gap. In the left sketch, the cross-lingual gaps are due to target bias, i.e., $\kappa=0$. In the right sketch, the cross-lingual gaps are due to variance, i.e., $\kappa=1$. We emphasize that $\kappa$ is deterministic for a given model and example pair. 
A key objective of our work is to find the range of $\pi$. The two mixture components have different expected behavior that allows us to establish the dominant component through a few targeted ablations, described in the rest of this section. Detailed proofs for all the subsequent results can be found in Appendix~\ref{sec:proofs}. 



\subsection{Decomposition of cross-lingual gaps into biased and unbiased components}
\label{sec:model:ensemble}
In this section, we study the nature of source-target gaps induced by the biased and unbiased errors. Specifically, we discuss how the likelihood of source-target agreement transforms with reduced response variance, i.e., increased value of $\alpha_s$ or $\alpha_t$.
\paragraph{Source-target gaps due to biased error ($\kappa=0$).}

\begin{prop}
 When the cross-lingual error is biased ($\arg\max \vec{z}_s \neq \arg\max \vec{z}_t$), the probability of a shared response, $\Pr(\hat{y}_s = \hat{y}_t)$, asymptotically approaches zero as response variance decreases (i.e., as the scale parameter $\alpha \to \infty$). 
\label{prop:1}
\end{prop}
The proof is based on the observation that as $\alpha_s\to\infty, \alpha_t\to\infty$, the sampling distributions are Dirac delta at their respective unequal modes.  

\paragraph{Source-target gaps due to unbiased error ($\kappa=1$).} 
\begin{prop}
When the cross-lingual error is unbiased, the probability of shared response between source and target increases with decreased response variance. 

\label{prop:2}
\end{prop}
Please find the proof in Appendix~\ref{sec:proof:prop2}.



\paragraph{Decoding the nature of gaps.}
From Propositions~\ref{prop:1},~\ref{prop:2}, we observe that the bias and variance components respond differently to response variance reduction. This also intuitively follows from the illustrative examples in Figure~\ref{fig:problem_illustrated} (a). Reducing the radii ($\sqrt{\text{variance}}$) will make the responses from source and target agree more often only when there are no biases. 

    

The simplest way to control response variance is by altering the sampling temperature. In practice, however, we observed temperature altering introduced bias. For instance, with our running example: {\it When was Kreeda Bharti established?}, Gemini-2.5-Flash (with thinking) responded with 1986$\pm$0 with zero temperature but 1991.7$\pm$8.4 with default temperature. Recall the correct answer is 1992 and sampling with zero temperature although reduced variance introduced a non-zero bias. Another example was {\it In what year was Lake Fucino declared drained?} with correct answer being 1878. But the Gemini model responded with 1870$\pm$0 with temperature zero and 1875$\pm$5 with temperature 1. The temperature induced bias is likely related to out-of-distribution generalization error rooted in shifted distribution of preceding tokens~\citep{holtzman2019curious}.

We need an unbiased estimator that reduces response variance, and Monte-Carlo estimation is a natural choice.  
We elicit multiple responses for the same example, model and use the majority voted response from N responses~\citep{hastie2003trees}. We empirically validate that ensembling does indeed improve the overall source-target agreement in Section~\ref{sec:ensemble_expts}. Decreasing source-target gaps with reduced response variance (ensemble size) indicate that the cross-lingual gaps are due to unbiased error, i.e., $\pi>0.5$. We may estimate $\pi$ more accurately by estimating the fraction of examples on which ensembling reduced source-target gaps, which we discuss in more detail in Section~\ref{sec:ensemble_expts}.


   
\vspace{-2mm}

\subsection{Additional implications of unbiased error in target responses}
\label{sec:model:implications}
We derive a surprising implication when the cross-lingual gaps are due to unbiased errors. When using a majority-vote-estimator described in the previous section, the cross-lingual gaps must diminish when the model is sufficiently confident (low variance) in the source language. On the other hand, we cannot claim the same when the cross-lingual gap is due to bias. We further validate that cross-lingual error is unbiased by empirically verifying the analytical predictions from this section.


We begin by showing that the response variance in source and target are related. We refer to the probability of a sample matching the mode as confidence and represent variance as 1 minus confidence. For instance, when the response variance is 0, all the sampled responses match the mode with probability 1.
Below, we relate confidence in target with confidence in source language. 

\begin{prop}
    Recall the categorical sampling process when the cross-lingual error is unbiased ($\kappa=1$). Let $\gamma = \alpha_t / \alpha_s \in (0, 1]$ represent the relative scale of the target distribution compared to the source. The confidence (probability of the mode) in the target language, $\Pr(\hat{y}_t = y^{\text{mode}})$, is bounded below by a strictly increasing function of the source confidence $\Pr(\hat{y}_s = y^{\text{mode}})$:
$$ \Pr(\hat{y}_t = y^{\text{mode}}) \ge \frac{1}{1 + (m-1)^{1-\gamma} \left( \frac{1 - \Pr(\hat{y}_s = y^{\text{mode}})}{\Pr(\hat{y}_s = y^{\text{mode}})} \right)^\gamma}, $$
where $m$ is the total number of normalized concepts in the response space.
\label{prop:src_tgt_conf_rel}
\end{prop}

Please find the proof in Appendix~\ref{sec:proof:prop3}. We highlight the following important implications of this result:
\begin{enumerate}[topsep=0pt,noitemsep,leftmargin=*]
    \item When the source confidence is high (i.e., $\Pr(\hat{y}_s = y^{\text{mode}}) \approx 1$), the right-hand side of the bound also approaches $1$. Therefore, the target confidence must also be high, even if the target distribution inherently suffers from higher variance.
    \item Since the lower bound on target confidence is strictly monotonic with respect to source confidence, we should observe increasing source-target agreement (a suppressed cross-lingual gap) as confidence in the source language increases.
\end{enumerate}
We empirically validate both these observations in Section~\ref{sec:results}.

\section{Experimental Setup}
\label{sec:expt_setup}
{\bf Datasets:} We experiment with two recent benchmarks: (1) \eclektic{}, (2) \mmlu{}. 
\begin{itemize}[noitemsep,leftmargin=*,labelsep=0.5em,topsep=0pt]
    \item {\bf \eclektic{}}~\citep{goldman2025eclektic} constitutes factoid questions sourced from Wikipedia pages that exist only in a single language. Questions from these single-language pages define the {\it source} split. Translations into any other language make up the {\it target} split. 
    \item {\bf \mmlu{}} builds on the popular multiple-choice MMLU dataset~\citep{mmlu_mixup} to introduce new examples that randomly mixes languages of the question and its answer options. The original questions define the {\it source} split, since examples with the same language for the question and all options are likely in-distribution with respect to pretraining. The mixed-language questions constitute the {\it target} split. 
\end{itemize} 
Both datasets are knowledge-intensive and require recall from entities in foreign languages/scripts. In Appendix~\ref{appendix:multiloko_res}, we also extend some of our results to the Multiloko dataset~\citep{hupkes2025multiloko}. 

\noindent
{\bf Languages:} \eclektic{},~\mmlu{} cover twelve and five languages, respectively. Please see Appendix~\ref{sec:dataset_details} for more details about both datasets. 

\noindent
{\bf Models:} We experiment with five SoTA LLMs. For closed-source LLMs, we chose from the Gemini~\citep{comanici2025gemini} series --- Gemini 2.5 Flash, Gemini 2.5 Pro --- and the GPT series~\citep{openai2024openai, openai2024gpt4technicalreport} --- GPT-5 mini, GPT-5. As a representative open model, we use Deepseek-R1~\citep{deepseekai2025deepseekr1incentivizingreasoningcapability}. 

\noindent
{\bf LLM-as-judge.} Responses in \eclektic{} are freeform text, which we rate using an LLM-judge. The LLM-judge determines if the model's response matches the reference; please refer to Appendix~\ref{sec:autochecker} for the LLM prompt. 
Using manual checks, we find that the LLM-judge is largely accurate on English instances but has slightly higher error rates on non-English instances. For a stronger validation of our claims without the influence of noise from the LLM-judge's ratings, we use a split of \eclektic{} that requires a year as an answer, which we refer to as~{\bf \neclektic{}}. Questions of the kind {\it ``In which/what year was $\dots$"} make up about 18\% of \eclektic{}. On \neclektic{}, we use a regex to extract years from the reference and response, and assess binary correctness (correct/incorrect) via an exact match. 


\section{Experimental Results}
\label{sec:results}

We empirically validate the following \textbf{two key findings} that shed new light on cross-lingual gaps:
\begin{enumerate}[noitemsep,leftmargin=*,labelsep=0.5em,topsep=0pt]
    \item Reducing response variance in \emph{both source and target languages} using simple inference-time ensembling techniques helps diminish cross-lingual gaps ($\S$\ref{sec:ensemble_expts}).
    \item Cross-lingual gaps reduce when response variance in the \emph{source language} reduces. In other words, higher source confidence leads to higher target confidence ($\S$\ref{sec:confidence_vs_gaps}).
    \looseness=-1
\end{enumerate}

\subsection{Cross-lingual gaps are largely due to variance}
\label{sec:ensemble_expts}

We sample ten responses per example for any LLM at default temperature with random seeds for each prompt. We measure the divergence between averaged target and source responses as the ensemble size increases from one to ten.%

\paragraph{\eclektic{}.}  In Figure~\ref{fig:point_robust}(a), we plot the average distance between source and target responses to the same question as a function of ensemble size. Since the responses in \eclektic{} are free-form text, we embed the responses using the \texttt{text-multilingual-embedding-002}~\citep{googleTextEmbeddings} model from Vertex AI~\citep{vertexai}. For each question, we compute the L2 distance between the average embedding of source responses and the average embedding of target responses from all target languages (i.e., across eleven languages other than the source). Finally, we compute an average L2 distance across all questions.
\begin{itemize}[itemsep=0pt, topsep=0pt, leftmargin=*]
\item {\bf Oracle.} We plot an ``oracle value'' in Figure~\ref{fig:point_robust}(a) using the average L2 distance across responses (to the same question) that are marked as correct by the LLM judge, thereby ensuring their equivalence. The oracle embedding distance is non-zero due to linguistic variations in the text. We also shows a confidence interval band around the oracle value using Bootstrap method, which captures the estimation error in the value.
\item {\bf $\pi$ estimate.} In each sub-plot, we present the value of $\pi=\mathbb{E}[\kappa]$ estimated as the fraction of examples where the L2 distance between source and target with ensemble size ten is within the confidence band of Oracle. That is, the fraction of examples where majority-voted response in source and target languages are distributionally indistinguishable according to our embeddings. 
\end{itemize}

\paragraph{\mmlu{}.} In Figure~\ref{fig:point_robust} (b), we plot the average Chi-squared distance (expression below) between the probability distributions over the four options in \mmlu{}.
\begin{align*}
    \text{Chi-squared}(p, q)=\sum_{x \text{ s.t. } p(x)+q(x)>0}\frac{(p(x)-q(x))^2}{p(x)+q(x)}
\end{align*}

\begin{itemize}[itemsep=0pt, topsep=0pt, leftmargin=*]
\item {\bf $\pi$ estimate.} In each sub-plot, $\pi$ is estimated as one minus the fraction of examples with mismatched predictions after ensembling. 
We approximate binary mismatch with a soft score (Appendix~\ref{sec:misc:mmlu_soft_approx}) using the difference of mode probabilities. 
\end{itemize}

\begin{figure}
    \centering
    \begin{subfigure}[t]{\linewidth}
        \includegraphics[width=\linewidth]{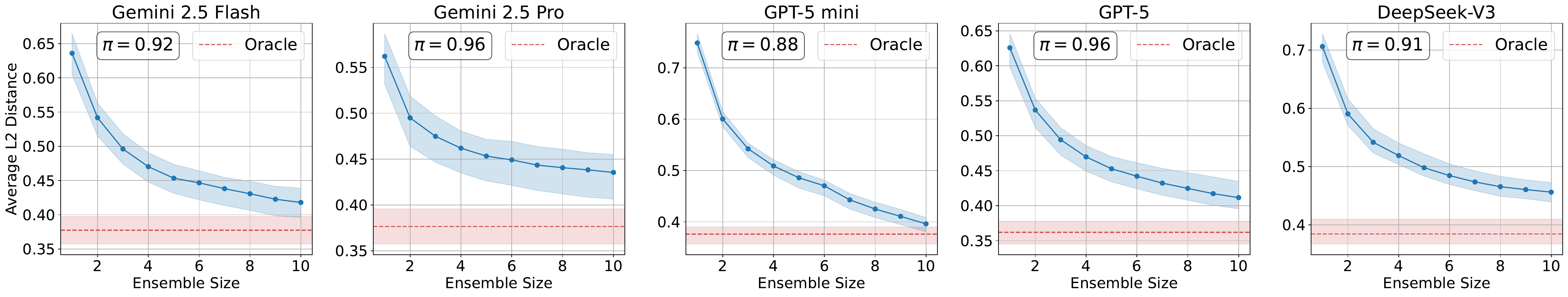}
        \caption{\eclektic{}.}
    \end{subfigure}

    \begin{subfigure}[t]{\linewidth}
        \includegraphics[width=\linewidth]{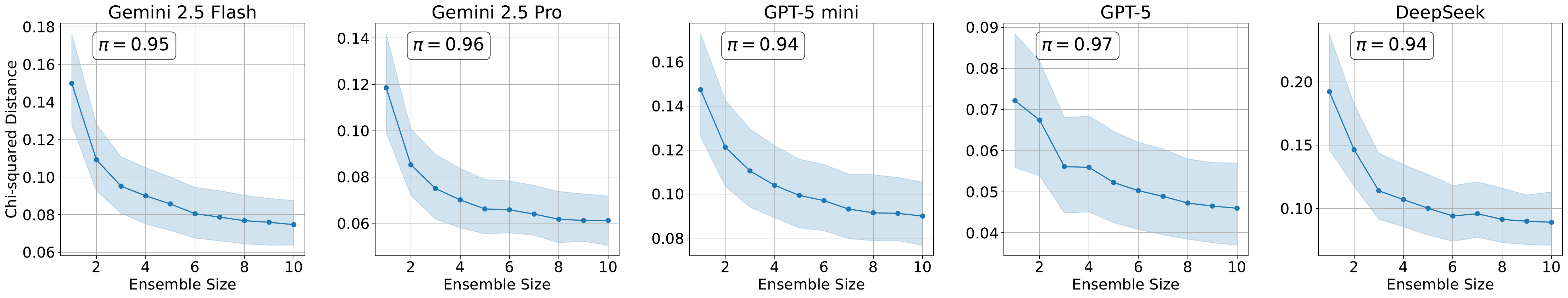}
        \caption{\mmlu{}.}
    \end{subfigure}
    \caption{Response ensembling from multiple forward passes gradually diminishes the source-target differences as illustrated on \eclektic{} (top) and \mmlu{} (bottom). Oracle value shown in red in (a) is the best expected value. Each plot shows an estimated value of $\pi$. Please refer Section~\ref{sec:ensemble_expts} for details.}
    \label{fig:point_robust}
\end{figure}

\paragraph{Observations.} We observe a steady decrease in source-target divergence with increasing ensemble sizes for both the benchmarks in Figure~\ref{fig:point_robust}. The estimated values of $\pi\approx0.9$ for \eclektic{} and $\pi\approx 0.95$ for \mmlu{} indicate that the noise is largely unbiased, i.e., $\kappa=1$ for 90-95\% of the examples. It is possible that the remaining examples are biased due to dataset errors. We report some spot-checks in Appendix~\ref{sec:misc:translation_errs}. In Appendix~\ref{sec:fine_ensemble_expts}, we analyze cross-lingual transfer at a finer level. In the Appendix section, and observe that the trend observed in this section holds for any combination high or low resource language transfer.

\subsection{Source variance drives target variance and cross-lingual gaps}
\label{sec:confidence_vs_gaps}
\begin{figure}[tb]
\begin{subfigure}[b]{\textwidth}
    \includegraphics[width=0.195\linewidth]{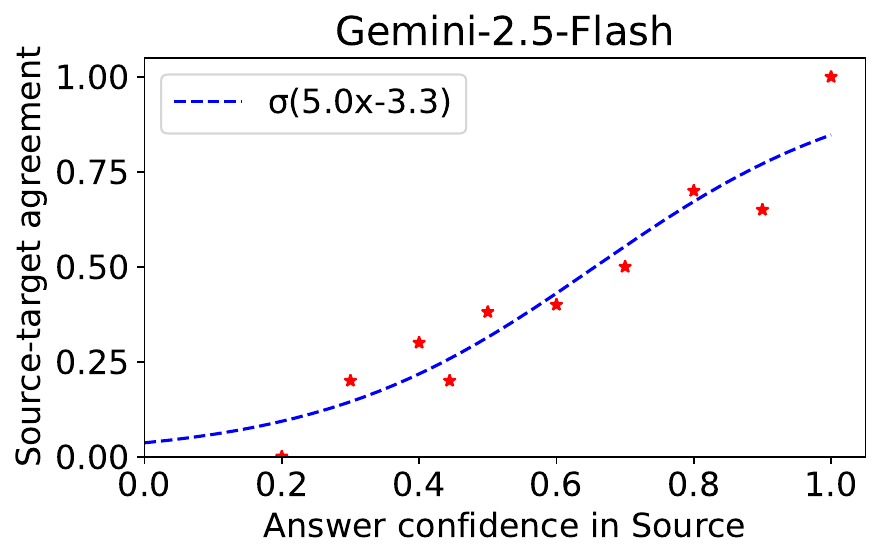}
    \includegraphics[width=0.195\linewidth]{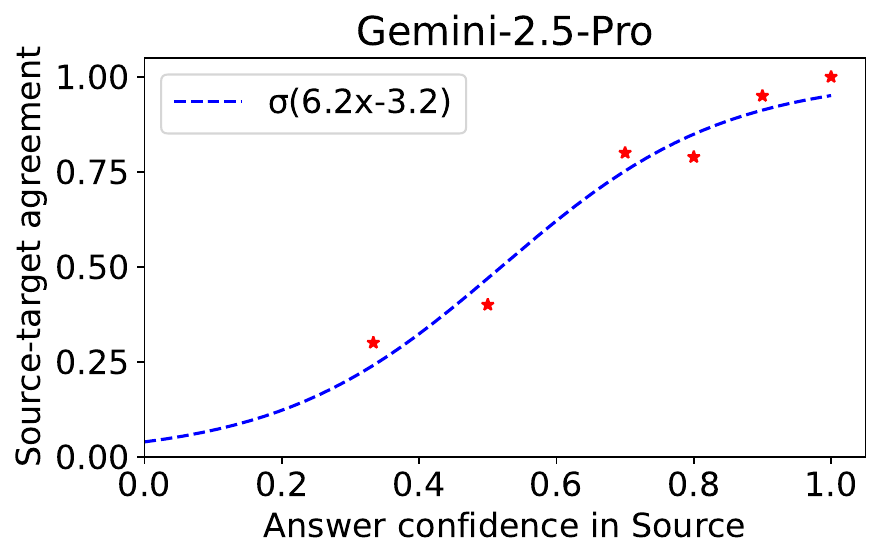}\hfill
    \includegraphics[width=0.195\linewidth]{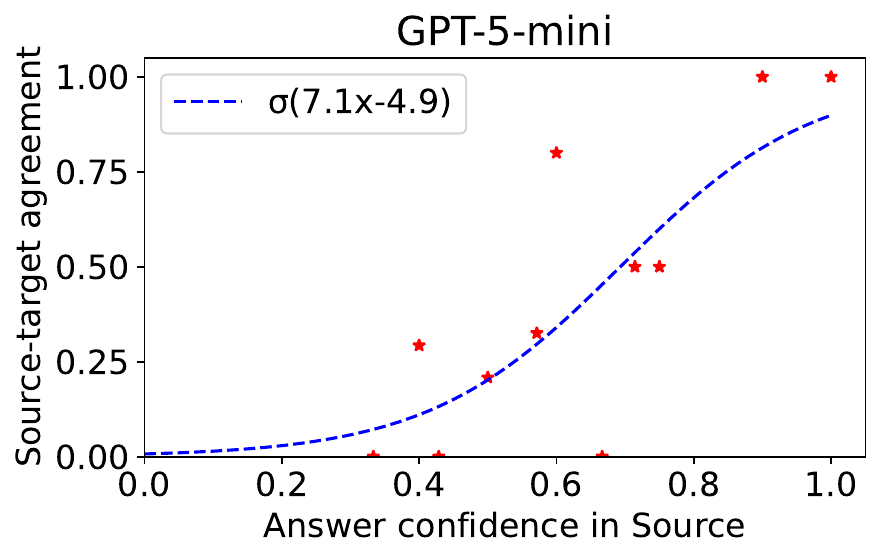}
    \includegraphics[width=0.195\linewidth]{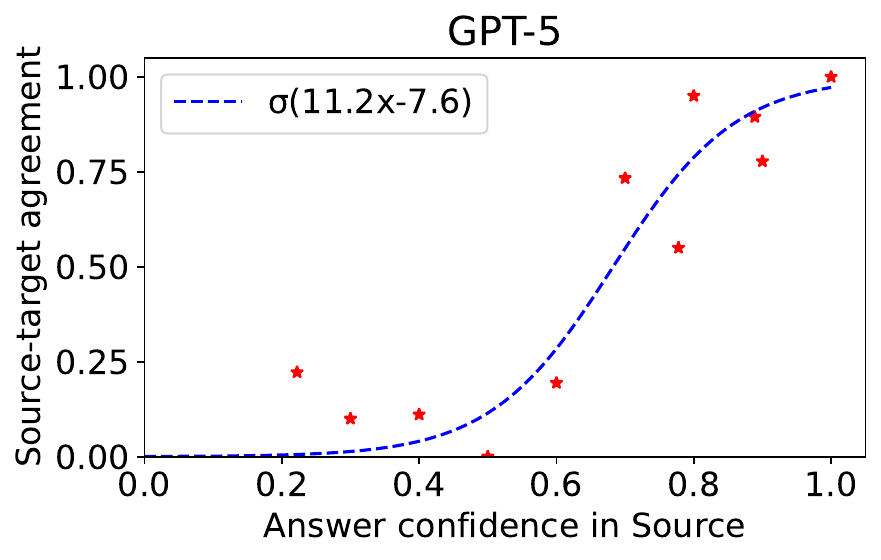}
    \includegraphics[width=0.195\linewidth]{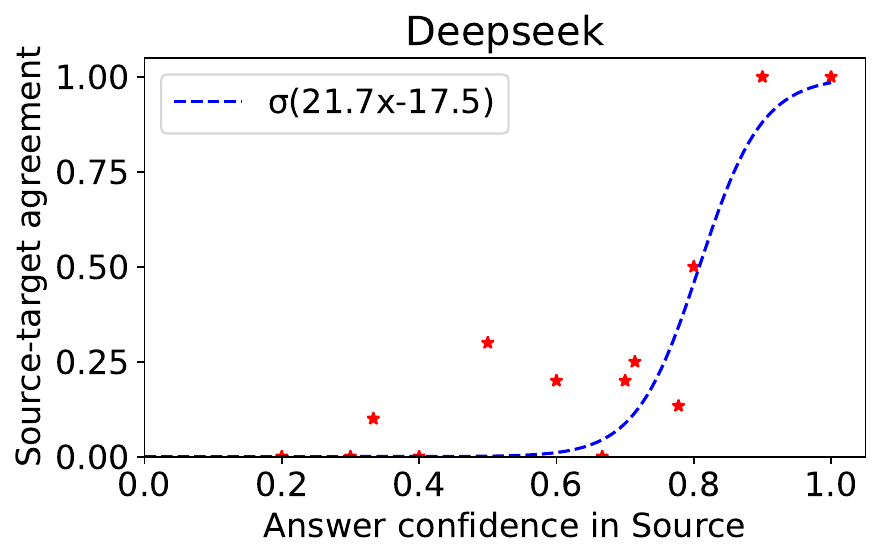}
    \caption{Year-\eclektic{}.}
\end{subfigure}
\begin{subfigure}{\textwidth}
    \includegraphics[width=0.195\linewidth]{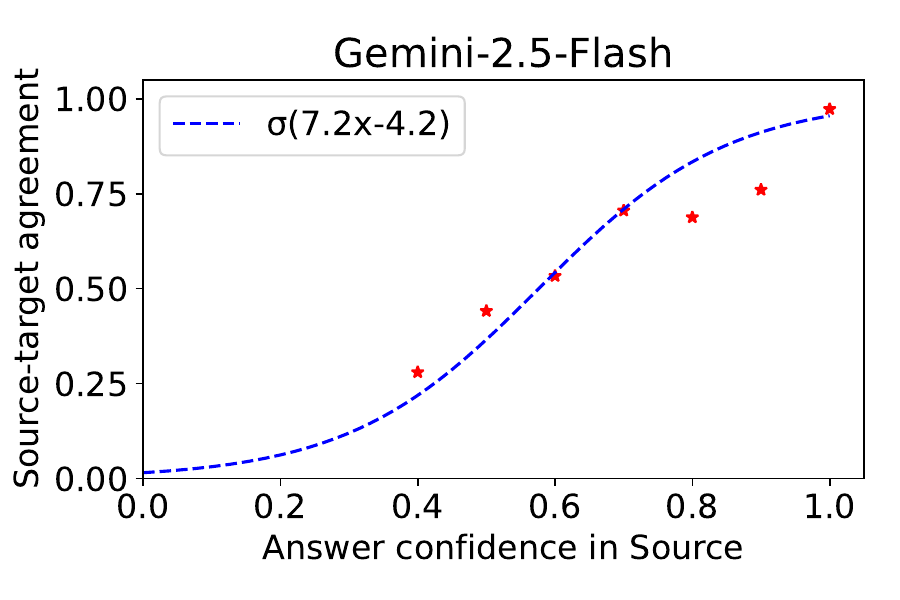}
    \includegraphics[width=0.195\linewidth]{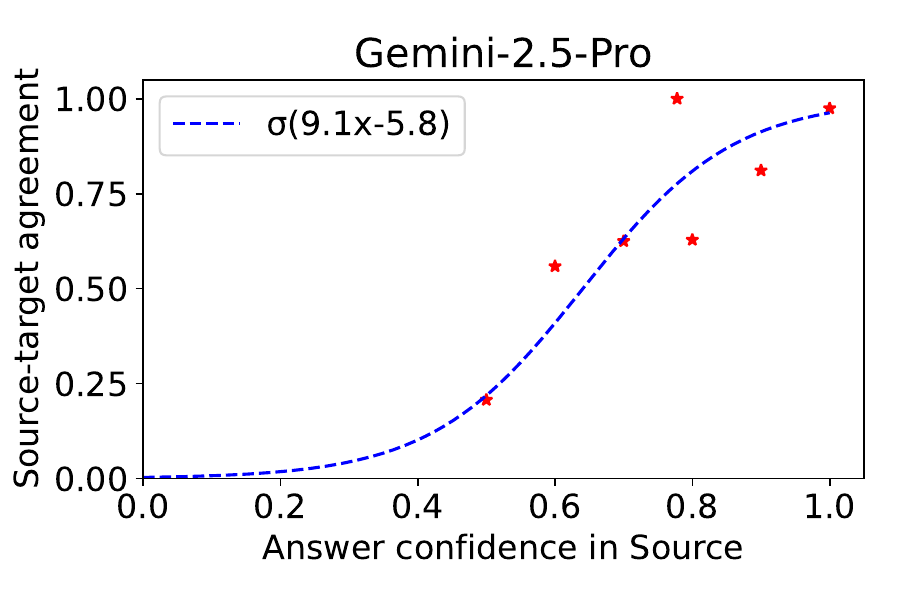}
    \includegraphics[width=0.195\linewidth]{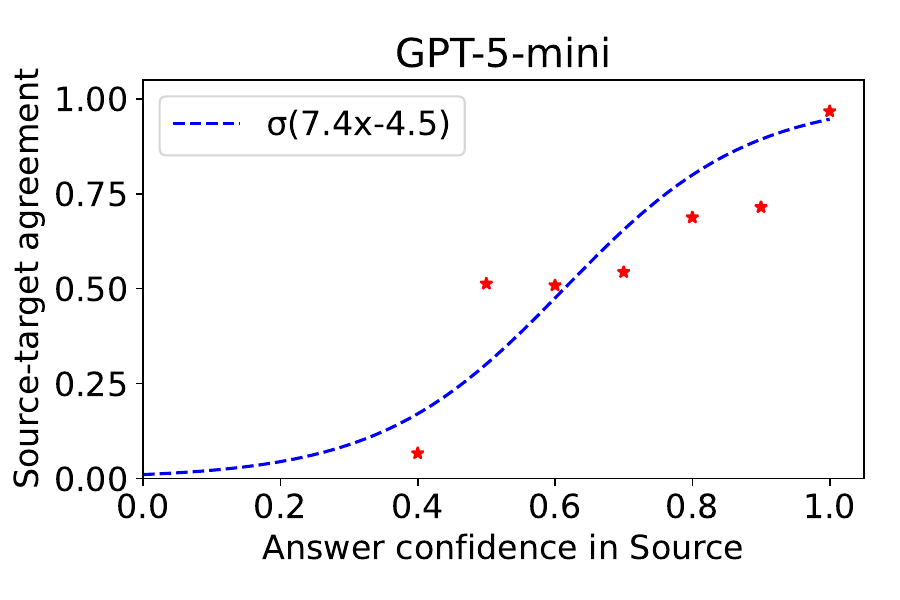}
    \includegraphics[width=0.195\linewidth]{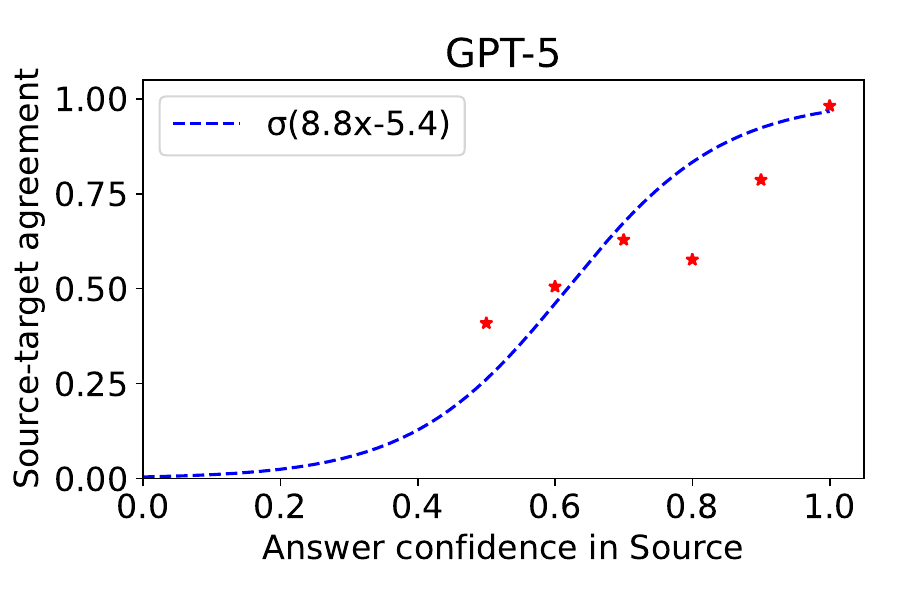}
    \includegraphics[width=0.195\linewidth]{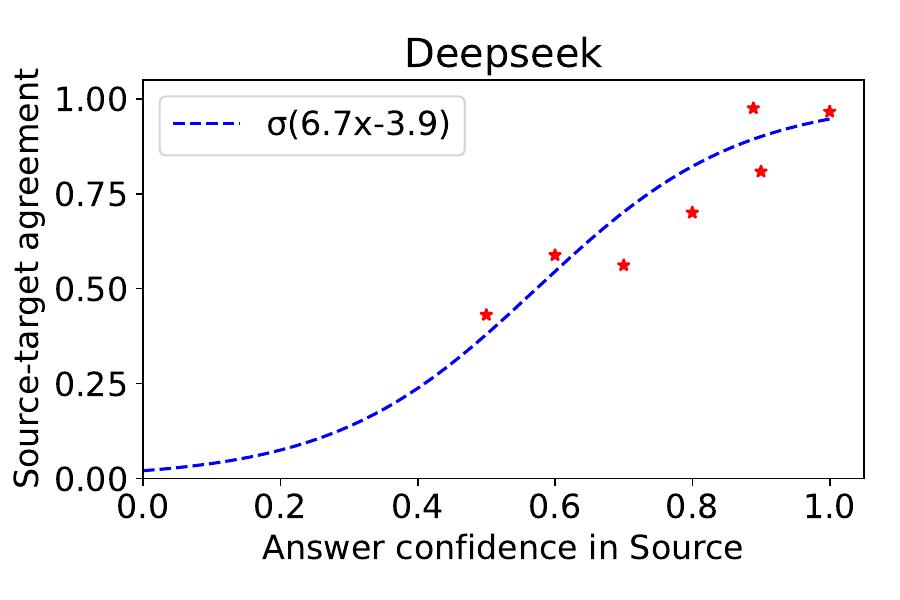}
    \caption{\mmlu{}.}
\end{subfigure}
\caption{Cross-lingual gaps diminishes with reduced variance in source language. Answer confidence is defined in Section~\ref{sec:model:implications}. High confidence in source leads to high confidence in target (Proposition~\ref{prop:src_tgt_conf_rel}), which should lead to improved agreement if there is no source-to-target bias. Section~\ref{sec:confidence_vs_gaps}.}
\label{fig:conf_vs_agree}
\end{figure}

In Figure~\ref{fig:conf_vs_agree}, for both \neclektic{} and \mmlu{}, we plot the average source-target agreement as the confidence in source increases. 
To recall, confidence is the probability of the mode, $\Pr(y_{s}^{\text{mode}})$ (as defined in Section~\ref{sec:model:implications} and Proposition~\ref{prop:src_tgt_conf_rel}). We empirically estimate confidence as the relative frequency of the mode across ten responses. 
We normalize the raw responses with \neclektic{} into the first year mentioned in the response in roman numerals through regex matching. 
We then compute for each question and (source, target) language pair, a binary score indicating if the majority-voted response match between the two languages: source, target.  
Finally, for both the datasets we estimate the source-target agreement by averaging across all the binary scores.

We observe consistent improvements in cross-lingual agreement as the source confidence improves, thereby confirming the claims in Section~\ref{sec:model:implications} and Proposition~\ref{prop:src_tgt_conf_rel}. We observe that the result is further validation of unbiased error because low response variance (high confidence) in source and target language leads to near-zero cross-lingual gaps.

\subsection{Other ensembling techniques}
\label{sec:input_ensemble} 
In  Section~\ref{sec:ensemble_expts}, we considered ensembling multiple responses by repeatedly prompting the same question. In this section, we present an alternate technique of reducing response variance: input ensembling. We elicit single response for multiple translations of the same prompt. 
Ensembling with semantically similar inputs is also known as test-time augmentation and has been found to be effective for improving robustness and uncertainty in various tasks such as image classification and segmentation~\citep{shanmugam2021better,moshkov2020test,ayhan2018test,krizhevsky2012imagenet}. 

\begin{wrapfigure}{r}{0.4\textwidth}
    \begin{subfigure}[t]{0.38\textwidth}
        \includegraphics[width=\linewidth]{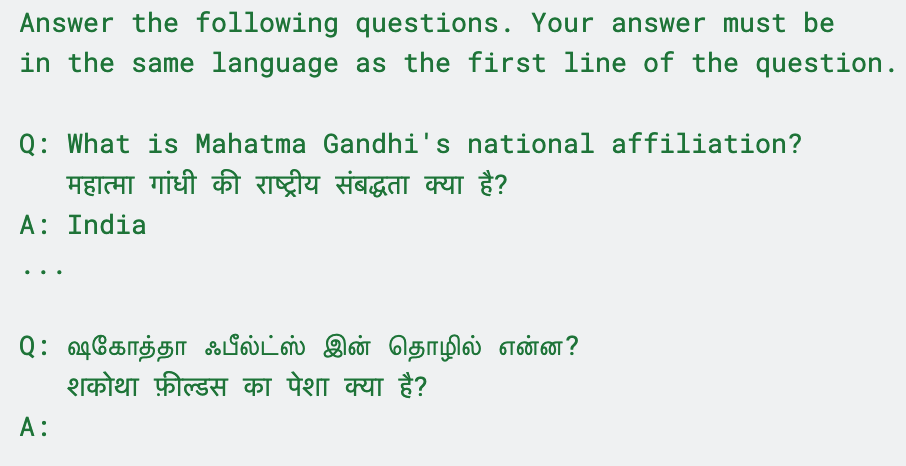} 
        \caption{Translation Ensemble (TrEn) with k=1.}
    \end{subfigure}
\end{wrapfigure}

We introduce an ablation called \textbf{Translation Ensemble (TrEn-k) }where we present the original question along with k translations and prompt for the answer. An example is shown in Figure to the right. In the appended translations, we ensure that we do not sample from the same script as the source. For example, if we are prompting a German-sourced question in Hindi, we do not sample translations from Hindi language or Latin script. This ensures that improvements with TrEn cannot simply be due to the accidental injection of the source question as a hint. We report results for k $\in \{1, 3, 5\}$.

We also include comparison with a popular baseline of Translating (the questions) to English and then Answer (in the original language), referred to as TEA~\citep{mmlu_mixup}.




\paragraph{Results.}
\begin{table}[htb]
    \centering
    \begin{tabular}{l|c|c|c|c|c}
    \toprule
    & G-2.5-Flash & G-2.5-Pro & GPT-5-mini & GPT-5 & Deepseek \\ 
    \midrule
         Baseline & 40.2 & 51.3 & 29.0 & 50.9 & 23.7 \\ 
    \midrule
         TEA & 41.9 & 54.7 & 27.9 & 50.1 &  36.2 \\
    \midrule
         TrEn-1 & 42.8 & 51.9 & 31.7 & 54.5 & 30.8  \\
         TrEn-3 & 44.7 & 50.5 & 33.7  & 54.8 & 35.8 \\
          \rowcolor{LightLimeGreen}
         TrEn-5 & 45.3 & 55.6 & 32.5 & 54.8 & 35.7 \\
    \bottomrule
    \end{tabular}
    \caption{Source-target transfer scores for \eclektic{}. Higher scores indicate better source-target transfer and better overall performance. TrEn-5 (highlighted) has consistently good performance. Please see Table~\ref{tab:results:inp_ens_acc} of Appendix for individual source-target accuracies.}
    \label{tab:results:input_ensemble}
\end{table}

Unlike the ensembling in Section~\ref{sec:input_ensemble}, TrEn yields only one response.  Therefore, we can directly evaluate for correctness and quantify transfer using transfer scores defined in~\citet{goldman2025eclektic} as summarized below:
\begin{align*}
    &A_{q,l}\triangleq 1\left(\text{ if the question $q$ is correct in both its source language and a target language } l\right)\\
    &\text{transfer-score}\coloneqq\mathbb{E}_{q,l}\left[A_{q,l}\right]
\end{align*}
Higher values of the score indicate better overall performance and transfer. The score is 100 only if source and target accuracies are perfect, and thus sub-perfect scores need not indicate high cross-lingual gaps. 

In Table~\ref{tab:results:input_ensemble}, we show transfer scores for our proposed input ensembling techniques using various models. We observe: (1) Consistent improvements in transfer scores from TrEn-1 to TrEn-5. (2) TEA improved over baseline only 2/5 times. (3) TrEn-5 had relative improvements over baseline of 8-13\% for Gemini/GPT class of models and 50\% on Deepseek. 

TrEn-5 uses oracle translations of the prompts that are readily available in the dataset, but are inaccessible in practice. Despite its practical limitations, effectiveness of TrEn contributes corroborating evidence to our main claim that cross-lingual error is unbiased. We observe that TEA, TrEn are approaches known to the community~\citep{mu2024revealing,zhou2023enhancing,qin2023cross,wang2025calm}, our analysis only further explains and validates their effectiveness with contemporary LLM models. 

We skip reporting results on \mmlu{} because each question in the dataset contains multiple languages. As a result, the dataset is not suitable for monolingual translations. Instead, we report results on the Multiloko dataset (described in Appendix~\ref{sec:dataset_details}) in Appendix~\ref{appendix:multiloko_res}. Our observed trends from Table~\ref{tab:results:input_ensemble} generalize well to Multiloko as well.

\section{Related Work}
{\bf Multilingual LLMs.} Prior work has focused on understanding LLMs' surprisingly strong performance on sparsely represented languages. \citet{brinkmann2025large} argued the existence of sparse multilingual features that bridge languages. \citet{schafer2024role} showed in a toy setting that imbalanced language representation during training is necessary for comparable multilingual performance. Models trained on largely imbalanced corpora (e.g., Llama-3, Gemini) often match or outperform models trained on more balanced multilingual datasets (e.g., Aya 23~\citep{dumas2024llamas}).

{\bf Unsupervised Domain Adaptation (UDA).} Cross-lingual gaps are a special case of UDA where the domain is defined by the language. UDA has a very rich literature~\citep{kouw2019review} where a common approach is to augment the objective to reduce source-target divergence in representations along with a predictive loss on the source. Methods differed in their choice of divergence measures and optimization~\citep{ganin2016domain}. The classic work of~\citet{ben2006analysis} gave an upper bound on target risk that depends on source risk and source-target representational divergence. Thereby, many approaches suggested expensive pretraining interventions to minimize the representational divergence, including multiple recent works~\citep{wang2024bridging,liu2025middle,ranaldi2023empowering,blum2025beyond}. This line of past work reflects on the traditional understanding that misaligned representations cause cross-lingual gaps. Our analysis attributed the gaps to variance and illustrated the promise of some simple approaches for fixing these gaps. Our analysis/findings strike a chord with self-consistency~\citep{wang2023selfconsistencyimproveschainthought} approach of reducing variance and improving performance in reasoning models on tasks that require thinking.

{\bf Representation-level interpretability analysis.}
Past work on interpreting language fragility in LLMs are also relevant to our work. 
\citet{fierro2024multilingual,wang2025lost} argue that LLMs translate multilingual inputs to English in intermediate representations, obtain an answer and translate them back to the original language. \citet{wang2025lost} rationalized cross-lingual gaps through errors in translating from the intermediate English answer to the final answer. \citet{lu2025paths} further argued that there are errors in both the levels of translations and proposed to fix them through steering vectors.  
Few other papers on mechanisms of factual recall are also related~\citep{ferrando2024know,yuksekgonul2023attention, meng2022locating}. 


\section{Discussion}
\label{sec:conclusion}
We closely analyzed the nature of cross-lingual errors in LLMs with knowledge-intensive tasks. 
We argue that increased response variance in target languages is a key cause of cross-lingual gaps. We also validate our hypothesis through targeted inference-time techniques that reduce cross-lingual gaps by simply ensembling multiple responses. We note that biases could arise from dataset errors. However, the datasets under consideration have been carefully vetted by human experts and are marked as having negligible human error. 

Overall, we hope that an improved understanding of the causes of cross-lingual gaps can help guide mitigation efforts (e.g., post-training to alleviate the issue of high response variance in the target language).

\noindent
{\bf Future Work and limitations.} (1) We proposed some inference-time mitigation strategies that are simple and effective, and leave mitigation via training for future work. (2) Our insights may also explain cross-modal gaps, i.e., performance disparities between text as input vs audio as input. We leave such generalizations for future work. 
(3) Our analysis may only apply to the languages covered by our datasets and to languages sufficiently well-represented in the training data, and will not apply to unseen (by LLM) languages. (4) We demonstrate that variance of responses increases in the target language, though the underlying cause remains unclear (Appendix~\ref{appendix:sbet}). Is increased variance in target a coping mechanism of LLMs to prevent perplexity loss from exploding in the presence of cross-language factual inconsistencies in pretraining data? We leave such analysis also for future work. \looseness=-1 
\bibliography{main}
\bibliographystyle{tmlr}

\clearpage
\appendix
\section{What determines the variance of responses?}
\label{appendix:sbet}
The main paper argued that cross-lingual gaps are due to high variance in source, which also determines the variance in target. The factors contributing to variance in source are unclear. We present some additional related insights in this section.

\paragraph{Entities are a hot-spot of cross-lingual gaps.}
Figure~\ref{fig:sbet} presents the results on \eclektic{} along with an illustration of our source borrowed entities in target (SBET) transformation. SBET bridged 60-70\% of  the cross-lingual gaps to the extent that the gaps between source and SBET are statistically insignificant (p=0.05). 

\begin{figure}[htb]
    \centering
    \includegraphics[width=0.40\linewidth]{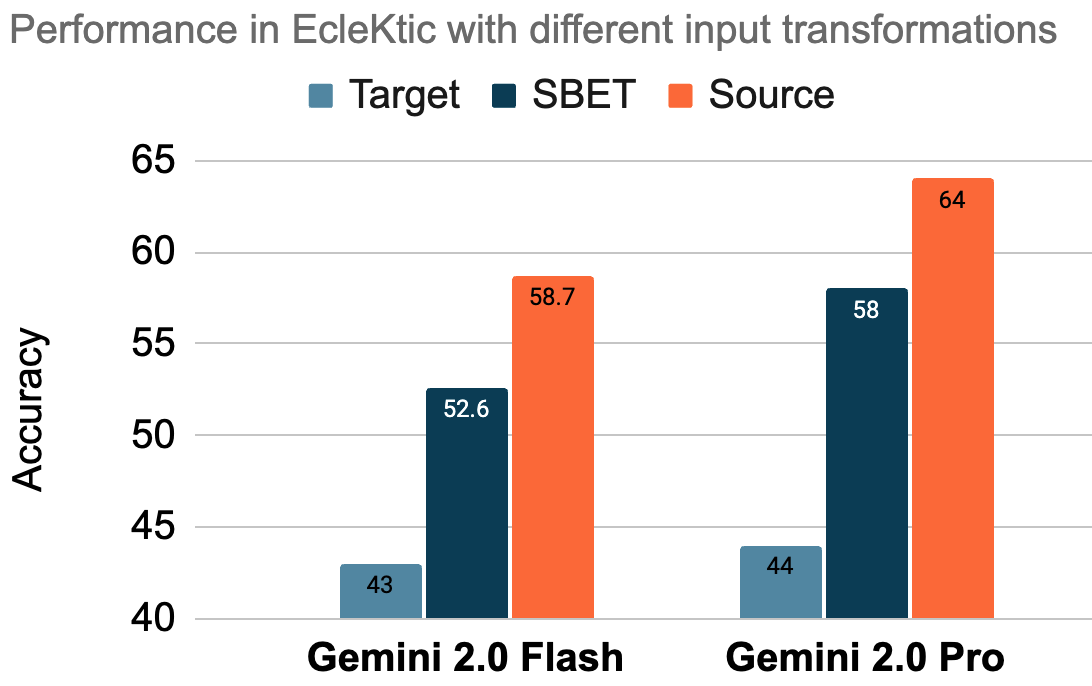}
    \includegraphics[width=0.54\linewidth]{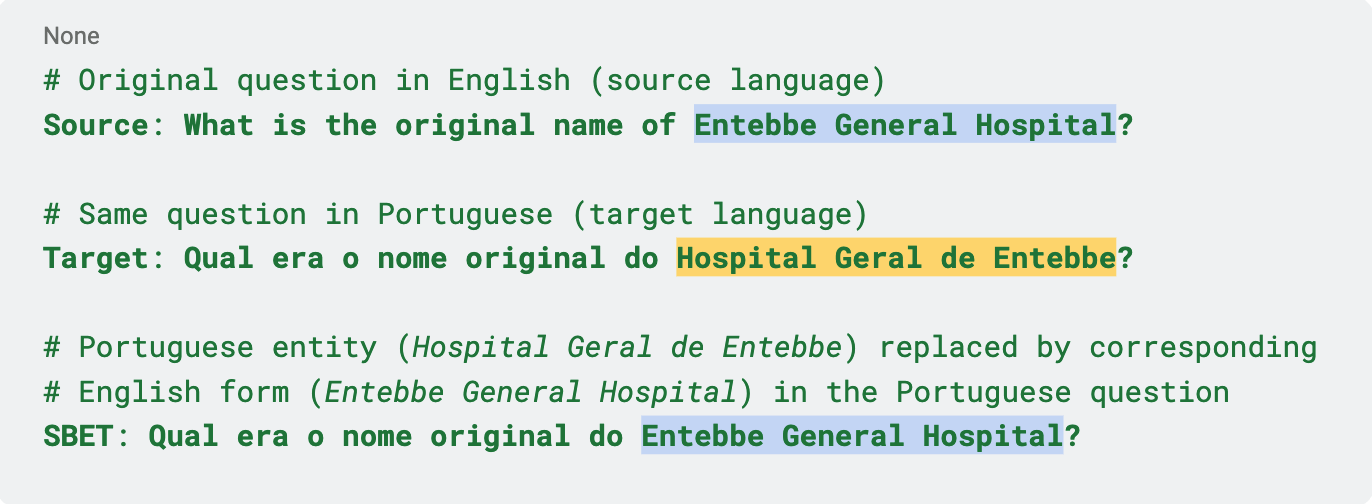}
    \caption{Illustration of Source Borrowed Entities in Target (SBET) and its performance. SBET transformation recovers 60\%-70\% of the cross-lingual drop suffered by Gemini 2.0 models when generalizing from Source to Target languages in \eclektic{}. Refer Section~\ref{appendix:sbet} for more details.}
\label{fig:sbet}
\end{figure}

\paragraph{(Multilingual) Popularity of Entities is uncorrelated with multilingual accuracy.}
\begin{wrapfigure}{r}{0.3\linewidth}
    \includegraphics[width=\linewidth]{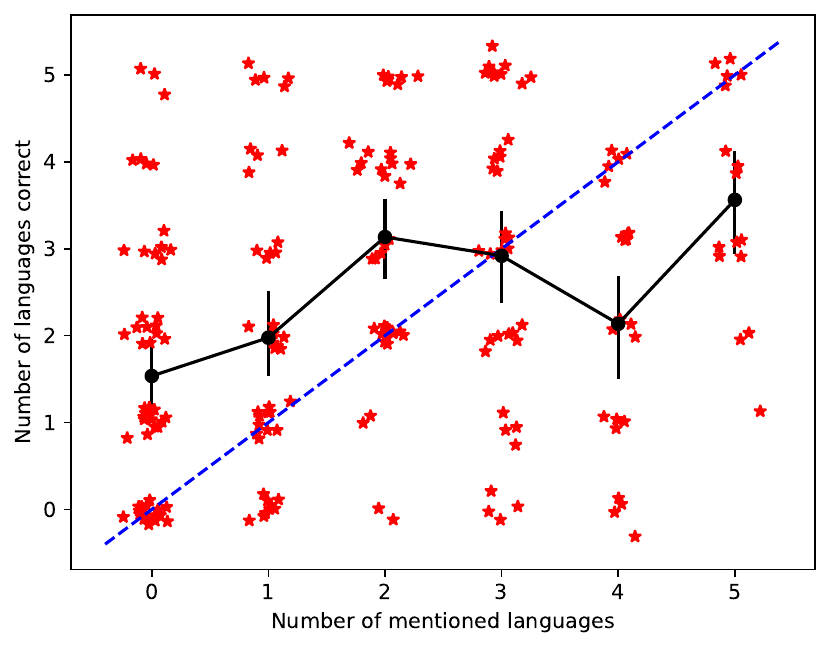}
    \vspace{-7mm}
\end{wrapfigure}
We may observe from SBET's effectiveness that the transfer of non-entity tokens is not a serious concern. 
It further demonstrated that entities are a hot-spot of cross-lingual factuality gaps. But it is unclear why entities in a target language are under-recognized irrespective of their mention statistics in the pretraining data. Take an example question from Hindi-sourced \eclektic{}: {\it Which body part of the Goddess fell at Pavagadh?} is correctly answered in almost every language even though the key entity ``Pavagadh'' is not mentioned in languages other than Hindi, English and Korean on the web. Therefore, entities are still recognized in foreign languages even though the exact surface form was never seen before. We also illustrate the poor correlation quantitatively on \eclektic{} in the right figure. We show the number of (target) languages an entity is mentioned (on the web) and the number of languages correctly answered on the vertical axis; we only considered languages with varying scripts for the analysis: {\it hi, zh, ko, ja, he, en}.  If not their multilingual multiplicity in pretraining data, what then determines confidence in source? Is knowledge consistency or duplication important?   

\section{How did we obtain embeddings of responses?}
\label{sec:embed_details}
For generating response embeddings we utilize the \textit{text-multilingual-embedding-002}~\citep{googleTextEmbeddings} API from the Vertex-AI platform. We generate embeddings for all the 10 responses for each question-language pair. Following that, we perform PCA analysis on the generated embeddings for a particular question ID for all languages. We found that often the first 2 PCA components were heavily influenced by the language script so we chose to plot 3rd and 4th component to convey semantic similarity. Finally we use kernel density plots with 5 levels to visualize the distribution or source and target language responses.

\section{Proofs}
\label{sec:proofs}

\subsection{Proof of Proposition~\ref{prop:2}}
\label{sec:proof:prop2}
We copy the statement of the proposition here for clarity. 

When the cross-lingual error is unbiased, the probability of shared response between source and target increases with decreased response variance.

\begin{proof}
We show that the probability of shared response monotonically increases with the decrease in response variance. We introduce a parameter $\kappa$ to control the variance. We take the derivative of probability of shared response with respect to $\kappa$ and show that it is always positive. Thereby, showing that increase in kappa (i.e., reducing response variance) leads to increased probability of shared response. 

  Recall that we denote by $\vec{z}_s, \vec{z}_t$ the source and target logits respectively. In this proof, we will use the shortforms $\alpha_s\vec{z}_s\rightarrow \vec{s}, \alpha_t\vec{z}_t\rightarrow \vec{t}$.
  Owing to bias, we expect the best answers according to source and target logits differ, i.e., $\arg\max_k \vec{s}\neq \arg\max_k\vec{t}$. We further use the short forms to denote $\mathbb{E}_p[S]=\sum_k p_k\vec{s}_k$ and $\mathbb{E}_q[T]=\sum_k q_k\vec{t}_k$.

  We will consider a peakiness controlling parameter $\kappa$ that determines the scale of logits and as a result controls the response variance.
  The probabilities after the softmax transformation takes the following form.
  $$p_k(\kappa) = \frac{\exp(\kappa s_k)}{\sum_i \exp(\kappa s_i)} \quad \text{and} \quad q_k(\kappa) = \frac{\exp(\kappa t_k)}{\sum_i \exp(\kappa t_i)}$$

  The probability of shared response requires marginalising over the output space: $\Pr(\hat{y}_s=\hat{y}_t) = \sum_k p_k(\kappa)q_k(\kappa)$. 
  We need to show that $\Pr(\hat{y}_s=\hat{y}_t)$ monotonically decreases with $\kappa$. That is, $\frac{\partial \Pr(\hat{y}_s=\hat{y}_t)}{\partial \kappa} < 0$ everywhere.

  We can work out $\partial p_k(\kappa)/\partial \kappa$ as below.
  \begin{align*}
    \partial p_k(\kappa)/\partial \kappa &= \partial \left\{\exp\log p_k(\kappa)\right\}/\partial \kappa = p_k\partial \log p_k(\kappa)/\partial \kappa\\
                                         &= p_k(s_k - \frac{1}{\sum_i \exp(\kappa s_i)}\sum_i \exp(\kappa s_i)s_i)\\
                                         &= p_k(s_k-\mathbb{E}_p[S]).
  \end{align*}
  Similarly, for $\partial q_k(\kappa)/\partial \kappa$.
  
  Finally, we can express $\frac{\partial \Pr(\hat{y}_s=\hat{y}_t)}{\partial \kappa}$ as below.

  \begin{align*}
    \frac{\partial \Pr(\hat{y}_s=\hat{y}_t)}{\partial \kappa} &= \sum_k p_k(\kappa)\frac{\partial q_k(\kappa)}{\partial \kappa} + q_k(\kappa)\frac{\partial p_k(\kappa)}{\kappa}\\
    &= \sum_k q_k(\kappa)p_k(\kappa)(s_k - \mathbb{E}_p[S]) + p_k(\kappa)q_k(\kappa)(t_k - \mathbb{E}_q[T]\\
    &= \left\{\sum_{k}p_k(\kappa)q_k(\kappa)\right\}\mathbb{E}_{pq}[S - \mathbb{E}_p[S]] + \mathbb{E}_{pq}[T - \mathbb{E}_q[T]]
  \end{align*}


\begin{align*}
    \frac{\partial \Pr(\hat{y}_s=\hat{y}_t)}{\partial \kappa} 
    &= \sum_k p_k(\kappa)q_k(\kappa)(s_k - \mathbb{E}_p[S]) + p_k(\kappa)q_k(\kappa)(t_k - \mathbb{E}_q[T])\\
    &=Z(\mathbb{E}_{pq}[S]-\mathbb{E}_p[S] + \mathbb{E}_{pq}[T] - \mathbb{E}_q[T])\\
    \quad & \text{where }Z=\sum_k p_k(\kappa)q_k(\kappa).
\end{align*}

In the unbiased case, the logits are directly proportional and the two distributions are related as $q_k\propto p_k^\alpha$ for some constant $\alpha>0$.

We will proceed by showing that both the quantities contributing to the sum are positive: $\mathbb{E}_{pq}[S]-\mathbb{E}_{p}[S]\geq 0, \mathbb{E}_{pq}[T]-\mathbb{E}_{q}[S]\geq 0$. We will show $\mathbb{E}_{pq}[S]-\mathbb{E}_{p}[S]\geq 0$, and the proof follows symmetrically for the other quantity. 

We may write $\mathbb{E}_{pq}[S]$ as $\frac{1}{Z}\sum_k p_kp_k^\alpha s_k = \frac{1}{Z}\sum_k p_k^{(1+\alpha)} s_k$. In order to prove that the value is greater than $\sum_k p_ks_k$, it suffices to show that $f(a)=\mathbb{E}_{p^a}[S]$ is an increasing function in $a$ when $a\geq 1$.  

We take the derivative of $f(a)$ with respect to $a$ and show that it is positive.
\begin{align*}
    \frac{d f(a)}{da} &= \frac{d}{da}\left\{\frac{1}{\sum_k p_k^a}\sum_k p_k^a s_k\right\}\\
    &= \frac{1}{(\sum_k p_k^a)^2} \left\{ \left(\sum_k p_k^a\right)\left(\sum_k p_k^a \ln(p_k) s_k\right) - \left(\sum_k p_k^a s_k\right)\left(\sum_k p_k^a \ln(p_k)\right) \right\}\\
    &\propto\sum_{i,j}p_i^{a}p_j^{a}(\ln(p_i)s_i-\ln(p_j)s_i)\\
    &=\sum_{i,j, i<j}p_i^ap_j^{a}s_i(\ln(p_i)-\ln(p_j)) + p_i^{a}p_j^as_j(\ln(p_j)-\ln(p_i))\\
    &=\sum_{i,j, i<j}\underbrace{p_i^{a}p_j^{a}(\ln(p_i)-\ln(p_j))(s_i-s_j)}_{>0}\\
    & >0
\end{align*}
The last inequality holds because $s_i<s_j\implies p_i<p_j$ and $s_i>s_j\implies p_i>p_j$.

Since $f(a)$ monotonically increases with $a$. The expression of our interest: $\mathbb{E}_{pq}[S] = \mathbb{E}_{p^{1+\alpha}}[S]$ is greater than original expectation $\mathbb{E}_p[S]$ if $\alpha>0$. Similarly, we can show that $\mathbb{E}_{pq}[T]>\mathbb{E}_q[T]$. 

Both the inequalities finally prove that the derivative of shared response probability with $\kappa$ is positive.   
\end{proof}






\subsection{Proof of Proposition~\ref{prop:src_tgt_conf_rel}}
\label{sec:proof:prop3}
We copy the proposition statement here for ease of reading. 

Recall the categorical sampling process when the cross-lingual error is unbiased ($\kappa=1$). Let $\gamma = \alpha_t / \alpha_s \in (0, 1]$ represent the relative scale of the target distribution compared to the source. The confidence (probability of the mode) in the target language, $\Pr(\hat{y}_t = y^{\text{mode}})$, is bounded below by a strictly increasing function of the source confidence $\Pr(\hat{y}_s = y^{\text{mode}})$:
$$ \Pr(\hat{y}_t = y^{\text{mode}}) \ge \frac{1}{1 + (m-1)^{1-\gamma} \left( \frac{1 - \Pr(\hat{y}_s = y^{\text{mode}})}{\Pr(\hat{y}_s = y^{\text{mode}})} \right)^\gamma}, $$
where $m$ is the total number of normalized concepts in the response space.

\begin{proof}
Recall that in the unbiased case, the logit vectors in source and target are aligned: $\alpha_s\vec{z}_s$, $\alpha_t\vec{z}_s$. We will use the logit vector $\vec{z}$ in place of $\vec{z}_s$. Without loss of generality, we assume that the first index of logit vector is the mode, i.e., $\arg\max \vec{z}=1$.  

By definition, 
\begin{align*}
&\Pr(\hat{y}_s=y^{mode}) = \frac{1}{1+\sum_{i>1}\exp(\alpha_s\Delta_i)},\\
&\Pr(\hat{y}_t=y^{mode}) = \frac{1}{1+\sum_{i>1}\exp(\alpha_t\Delta_i)},\\
&\text{where }\Delta_i=\vec{z}[i]- \vec{z}[1].
\end{align*}

We focus on the denominator term $\sum_{i>1}\exp(\alpha_t\Delta_i)$ and derive the following:
\begin{align*}
    &\sum_{i>1}\exp(\dfrac{\alpha_t}{\alpha_s}\alpha_s\Delta_i) = \sum_{i>1}x_i^\gamma,\quad \text{where } \gamma=\dfrac{\alpha_t}{\alpha_s}, x_i = \exp(\alpha_s\Delta_i),\\
    &\sum_{i>1}x_i^\gamma \leq (m-1)\left\{\dfrac{\sum_{i>1}x_i}{m-1}\right\}^{\gamma}\quad \text{Due to Jensen's inequality on concave $x^\gamma$ for $\gamma\in (0, 1]$},\\
    &\implies \sum_{i>1}\exp(\alpha_t\Delta_i)\leq (m-1)^{1-\gamma}\left\{\sum_{i>1}\exp(\alpha_s\Delta_i)\right\}^\gamma
\end{align*}

We get the final bound shown in the proposition by observing that $$\sum_{i>1}\exp(\alpha_s\Delta_i) = \dfrac{1}{\Pr(\hat{y}_s=y^{mode})}-1.$$
\end{proof}

\section{Dataset details}
\label{sec:dataset_details}
    {\bf \eclektic{}}~\citep{goldman2025eclektic} dataset constitutes factual questions from Wikipedia pages that exist only in single language. Single language Wikipedia pages are used as a proxy for content that is unavailable or unpopular in other languages. Therefore, the benchmark proposed to validate the cross-lingual knowledge transfer on questions translated from the original language. Questions that are in the same language as the passage that contained the fact in pretraining data define the {\it source} split. While the questions in any other language make up the {\it target} split. It contains around 5500 examples and spans 12 languages.
    
    {\bf \mmlu{}}~\citep{mmlu_mixup} dataset alters MMLU~\citep{mmlu} to probe LLM's language generalization. MMLU is a multiple choice dataset spanning multiple tasks. \mmlu{} proposes to mixup the question by replacing the options with translations into random languages. Original questions make up the source split because examples with shared language for both question and options are likely in-distribution with pretraining. Likewise, questions with language mixed options make up the target split. We subsample around 2000 examples and this spans five languages. Figure~\ref{fig:mmlu_example} shows an example.



\begin{figure}[b]
    \centering
    \includegraphics[width=0.75\linewidth]{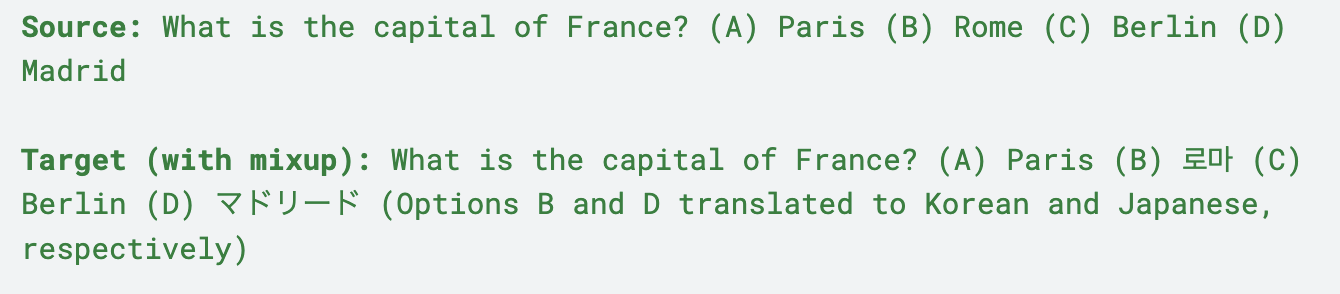}
    \caption{An example from the \mmlu{} dataset.}
\label{fig:mmlu_example}
\end{figure}

{\bf MultiLoKo}~\citep{hupkes2025multiloko} is a recently released benchmark for multilingual evaluation of LLMs covering 31 languages. MultiLoKo consists of three partitions: a main partition consisting of 500 questions per language, separately sourced to be locally relevant to
the specific language, and two translated partitions, containing human-authored
translations from 30 non-English languages to English and vice versa. For our use-case we use the \textit{dev-split} which is publicly available consisting of 250 questions per language along with human-authored translations.

\noindent
{\bf Languages:} \eclektic{} dataset spans twelve languages: German (de), Chinese (zh), Portuguese (pt), Spanish (es), Hindi (hi), Italian (it), Indonesian (id), Hebrew (he), Japanese (ja), French (fr), English (en), and Korean (ko). \mmlu{}, on the other hand, covers five languages: English (en), French (fr), German (de), Spanish (es), and Italian (it).

\section{Variance in Source determines the Variance in Target}
\label{sec:var_in_src_tgt}
In this section, we present results that support the text in Section~\ref{sec:confidence_vs_gaps}. 

Figure~\ref{fig:conf_vs_conf} shows average confidence in target with that of source for both the datasets.  
\begin{figure}[htb]
\begin{subfigure}[b]{\textwidth}
    \includegraphics[width=0.195\linewidth]{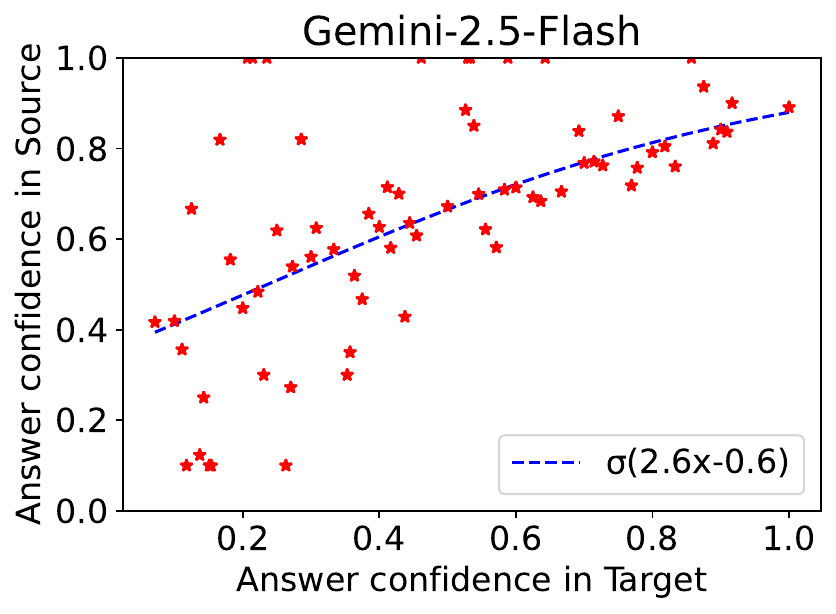}
    \includegraphics[width=0.195\linewidth]{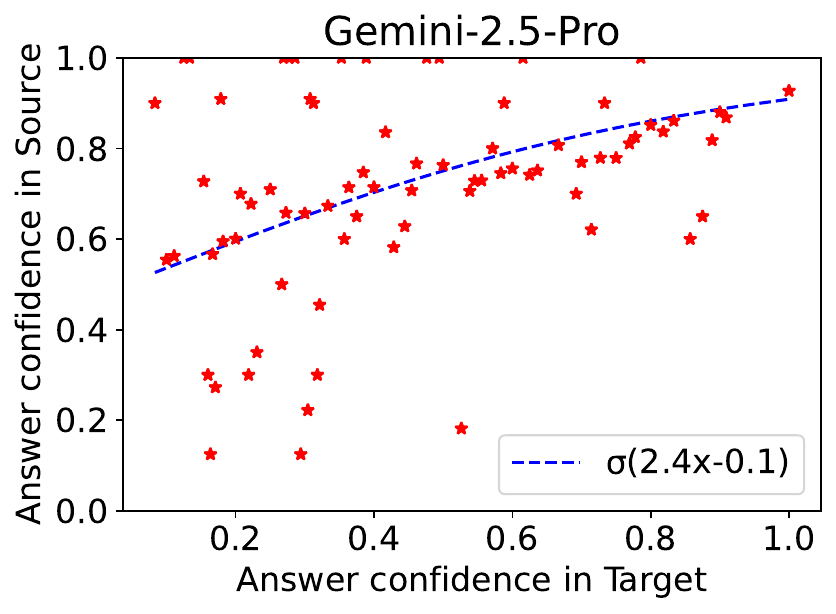}\hfill
    \includegraphics[width=0.195\linewidth]{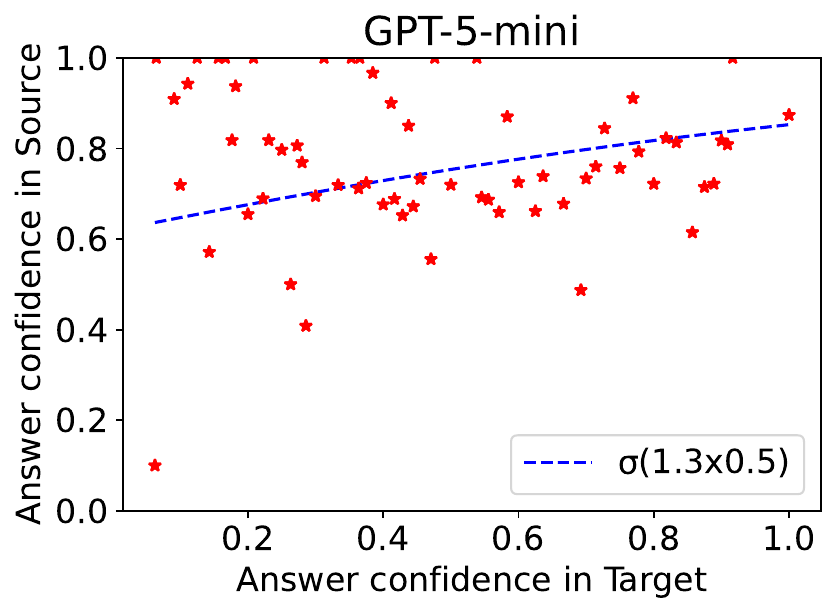}
    \includegraphics[width=0.195\linewidth]{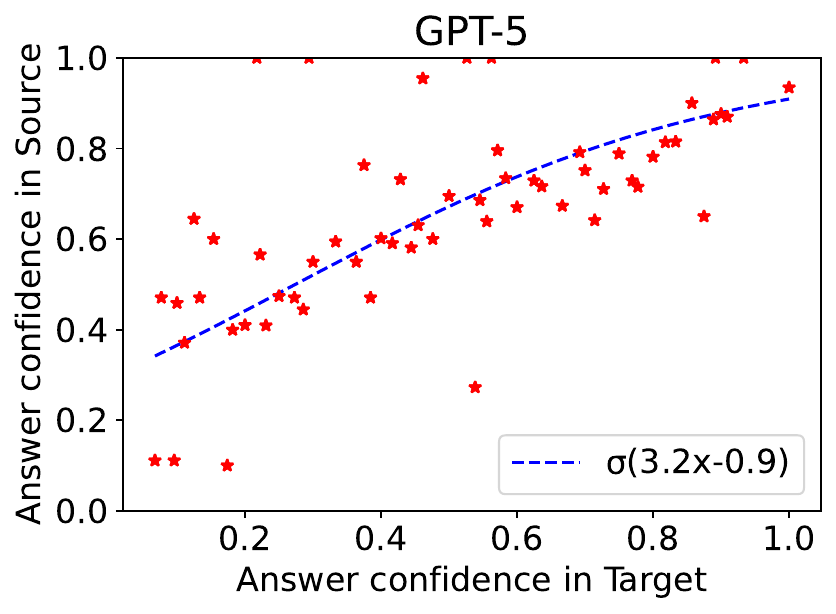}
    \includegraphics[width=0.195\linewidth]{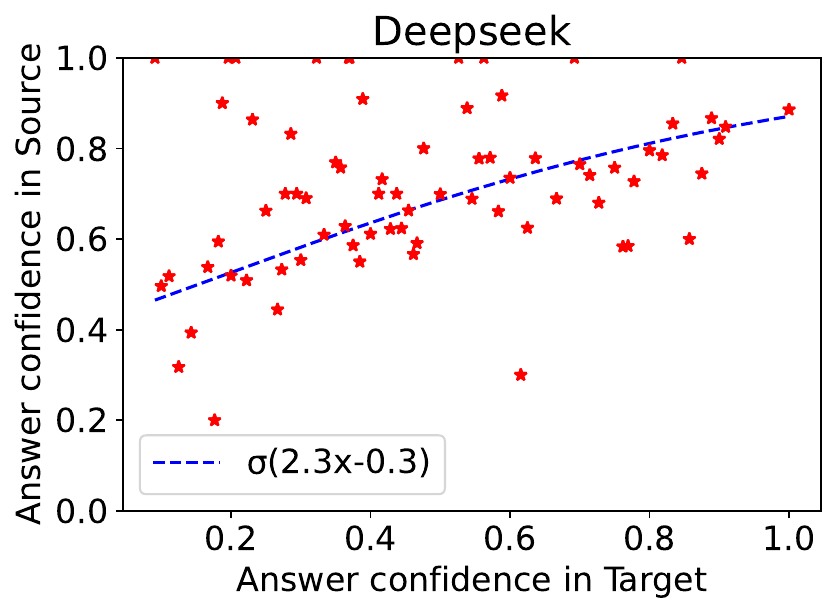}
    \caption{\eclektic{}.}
\end{subfigure}
\begin{subfigure}{\textwidth}
    \includegraphics[width=0.195\linewidth]{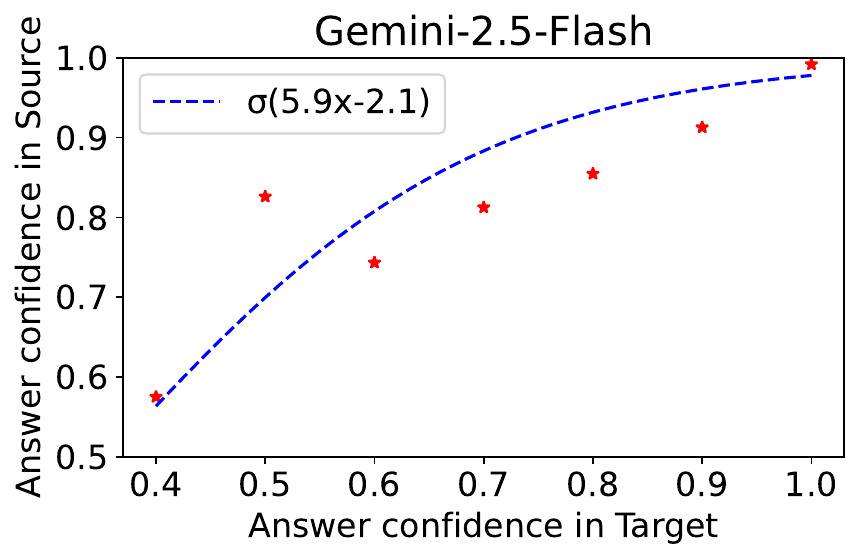}
    \includegraphics[width=0.195\linewidth]{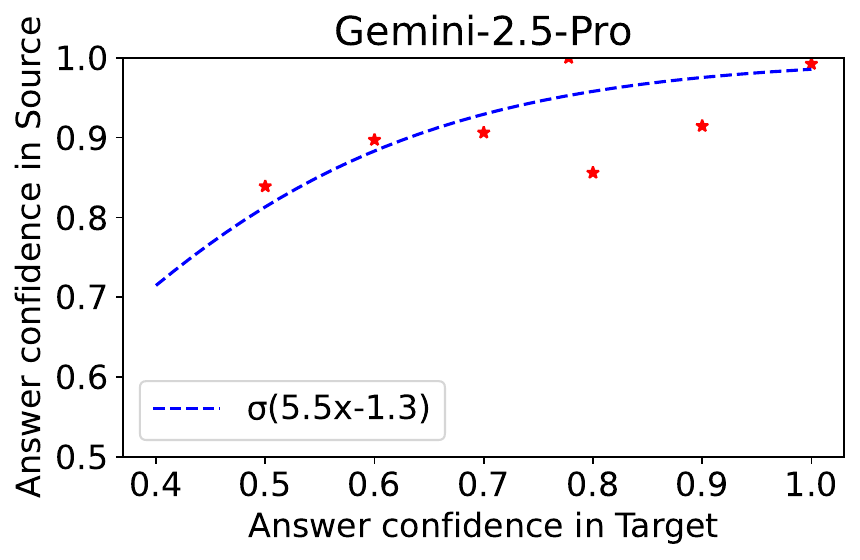}
    \includegraphics[width=0.195\linewidth]{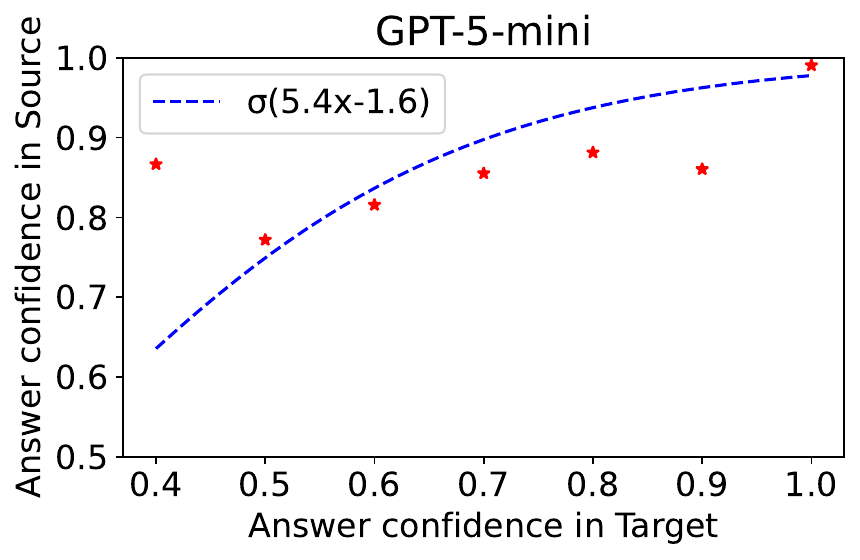}
    \includegraphics[width=0.195\linewidth]{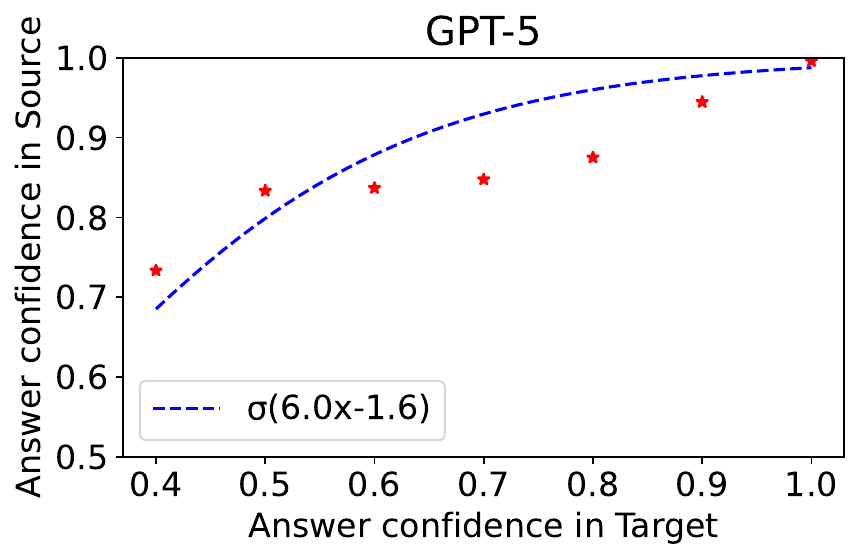}
    \includegraphics[width=0.195\linewidth]{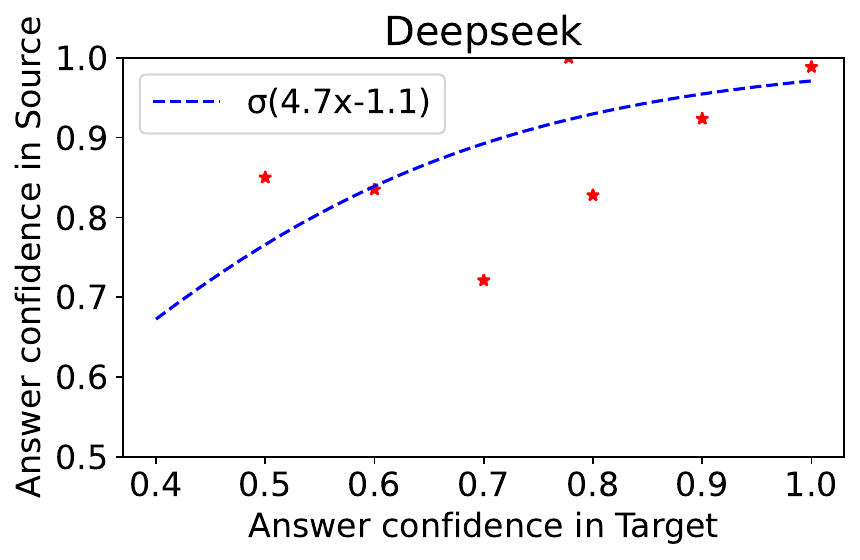}
    \caption{\mmlu{}.}
\end{subfigure}
\caption{Empirical validation that high confidence in source leads to high confidence in target (Proposition~\ref{prop:src_tgt_conf_rel}. Answer confidence is defined in Section~\ref{sec:model:implications}. The figure supports Section~\ref{sec:confidence_vs_gaps} of main text. Refer Section~\ref{sec:var_in_src_tgt} for more details.}
\label{fig:conf_vs_conf}
\end{figure}

\section{Results on \neclektic{}}
\label{sec:neclektic_results}
In Figure~\ref{fig:neclektic_acc}, we reproduce the results on \neclektic{} from Section~\ref{sec:ensemble_expts} and Figure~\ref{fig:point_robust}. We observe similar but cleaner trend on \neclektic{} of decreasing Mean Absolute Error (MAE) between the averaged source and target answers.  We also observe that the target accuracy increased with ensemble size such that the source-target accuracy differences at the extreme right are statistically insignificant for 3 of 4 models.

In Figure~\ref{fig:neclektic_conf_vs_gaps}, we reproduce the results on \neclektic{} from Section~\ref{sec:confidence_vs_gaps} and Figure~\ref{fig:conf_vs_agree}. The figure more cleanly demonstrates the trend of increasing agreement between source and target as the confidence in source increases. 

\begin{figure}
    \includegraphics[width=\linewidth]{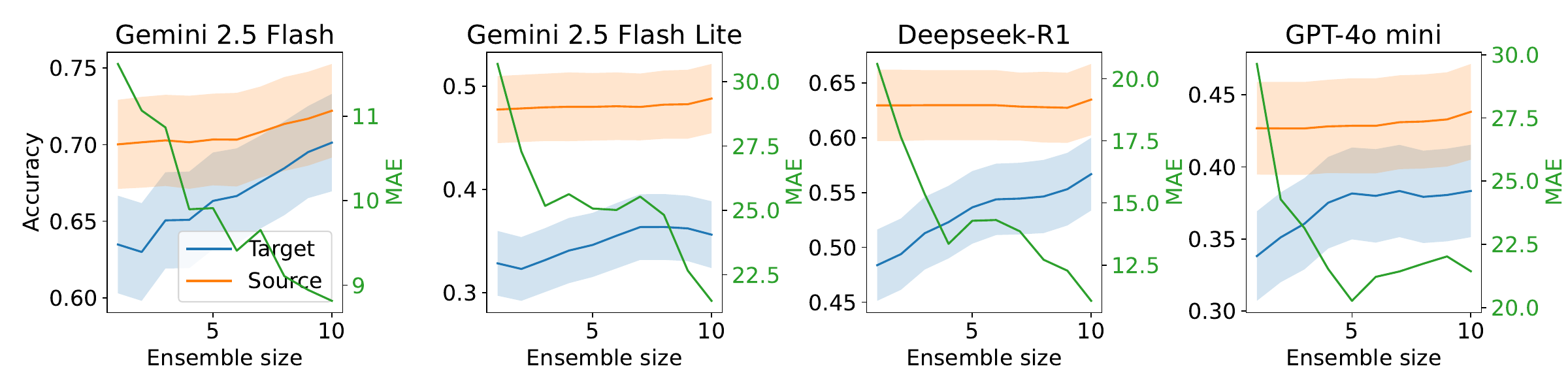}
    \caption{Reproducing results from Figure~\ref{fig:point_robust} on \neclektic{}.}
    \label{fig:neclektic_acc}
\end{figure}

\begin{figure}[htb]
    \includegraphics[width=0.245\linewidth]{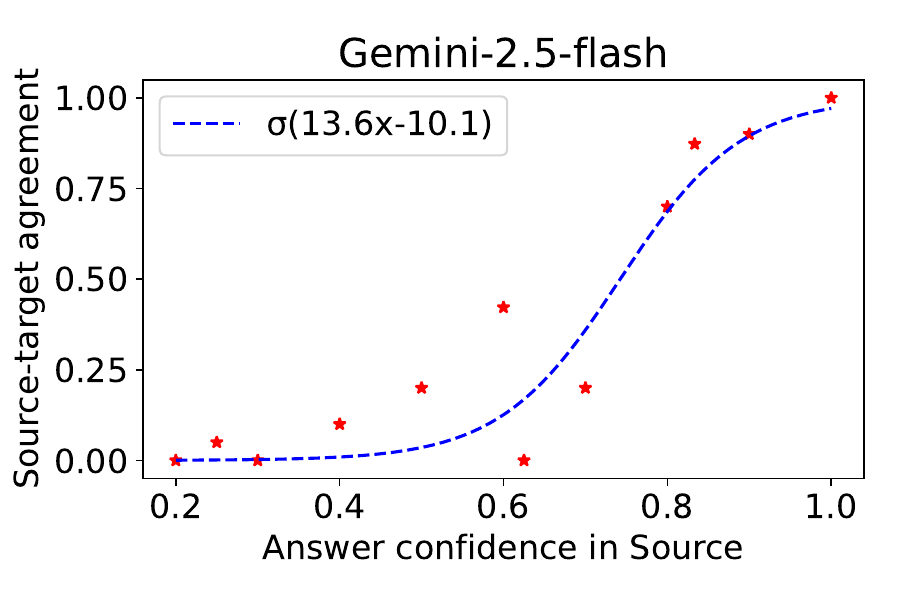}
    \includegraphics[width=0.245\linewidth]{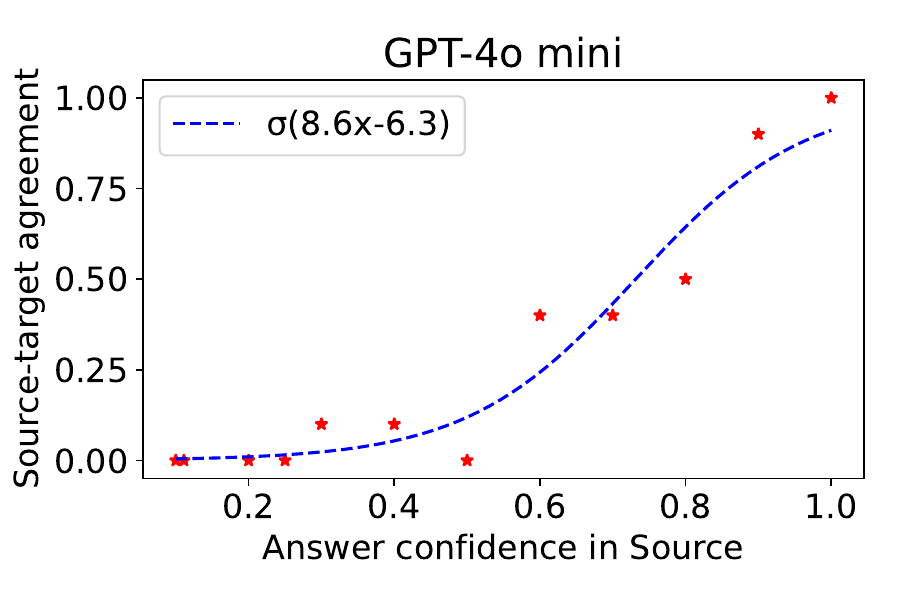}\hfill
    \includegraphics[width=0.245\linewidth]{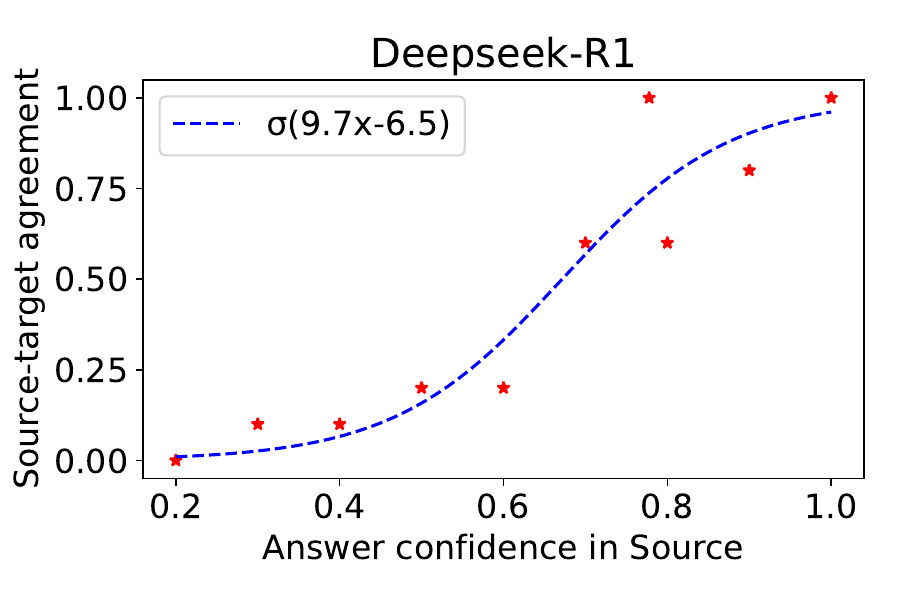}
    \includegraphics[width=0.245\linewidth]{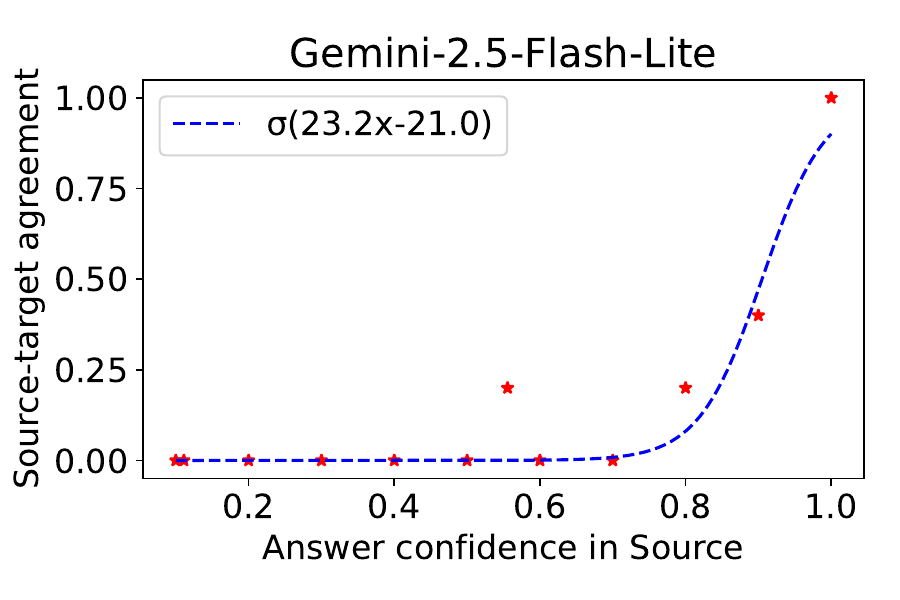}
    \caption{Reproducing results from Figure~\ref{fig:conf_vs_agree} on \neclektic{}.}
    \label{fig:neclektic_conf_vs_gaps}
\end{figure}

\section{Results on MultiLoKo}
\label{appendix:multiloko_res}
We also evaluate our Input Ensembling strategies as mentioned in Section~\ref{sec:input_ensemble} on MultiLoKo dataset~\cite{hupkes2025multiloko}. Since our evaluation strategy required multiple translations of the question in up to 5 languages, we decided to use only the English division of the MultiLoKo dataset as it contains translations in 30 languages, while for other languages translations are only available in English language which restricts their evaluation. As observed in Table~\ref{tab:results:multiloko}, following the trend as seen on \eclektic{} dataset in Table~\ref{tab:results:input_ensemble}, we observe improvements from TrEn-1 to TrEn-5. And TrEn-5 gives the best performance for most models. We noticed that some of these artifacts can also be influenced due to the input system prompt given to a specific model, hence the noise in trends.
\begin{table}[htb]
    \centering
    \begin{tabular}{l|c|c|c|c|c}
    & G-2.5-Flash & G-2.5-Pro & GPT-5-mini & GPT-5 & Deepseek \\\hline
         Baseline & 54.3 & 60.1 & 48.4 & 65.2 & 29.8 \\\hline
         TrEn-1 & 55.5 & 62.7 & 51.4 & 68.7 & 32.8 \\
         TrEn-3 & 56.3 & 61.8 &  50.2 & 67.1 & 32.7\\
         \rowcolor{LightLimeGreen}
         TrEn-5 & 60.0 & 63.3 & 49.4 & 68.5 & 33.3\\\hline
    \end{tabular}
    \caption{Source-Target transfer scores for MultiLoKo. Higher score indicate better Source-Target transfer and better overall performance. TrEn-5 (highlighted) has consistently good performance. Refer Section~\ref{appendix:multiloko_res}.}
    \label{tab:results:multiloko}
\end{table}

\begin{figure}[htb] 
    \centering 

    \begin{subfigure}[t]{0.32\linewidth}
        \includegraphics[width=\linewidth]{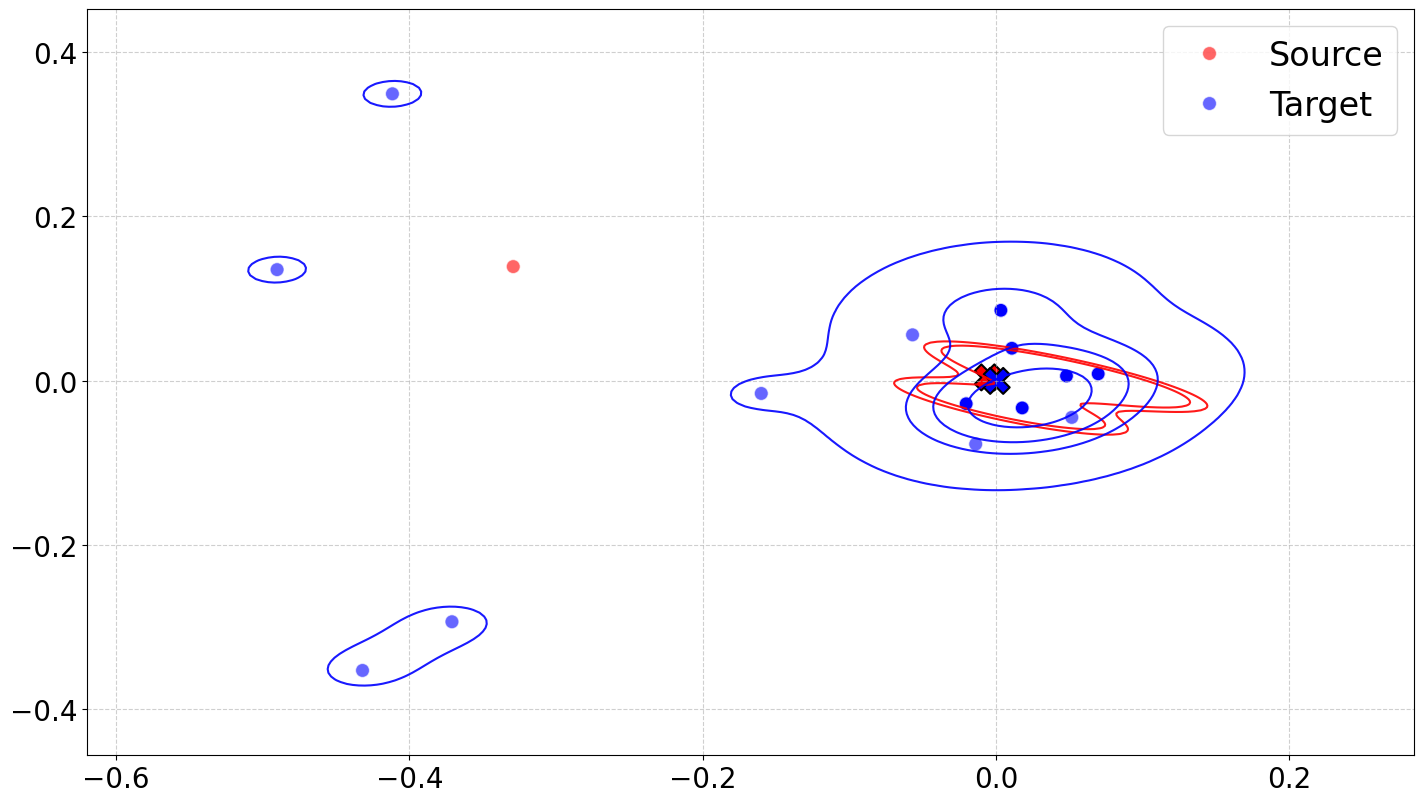}
        \caption{According to Benjamin Mazar and Yohanan Aharoni, to which tribe did the city of Dor belong?}
    \end{subfigure}
    \begin{subfigure}[t]{0.32\linewidth}
        \includegraphics[width=\linewidth]{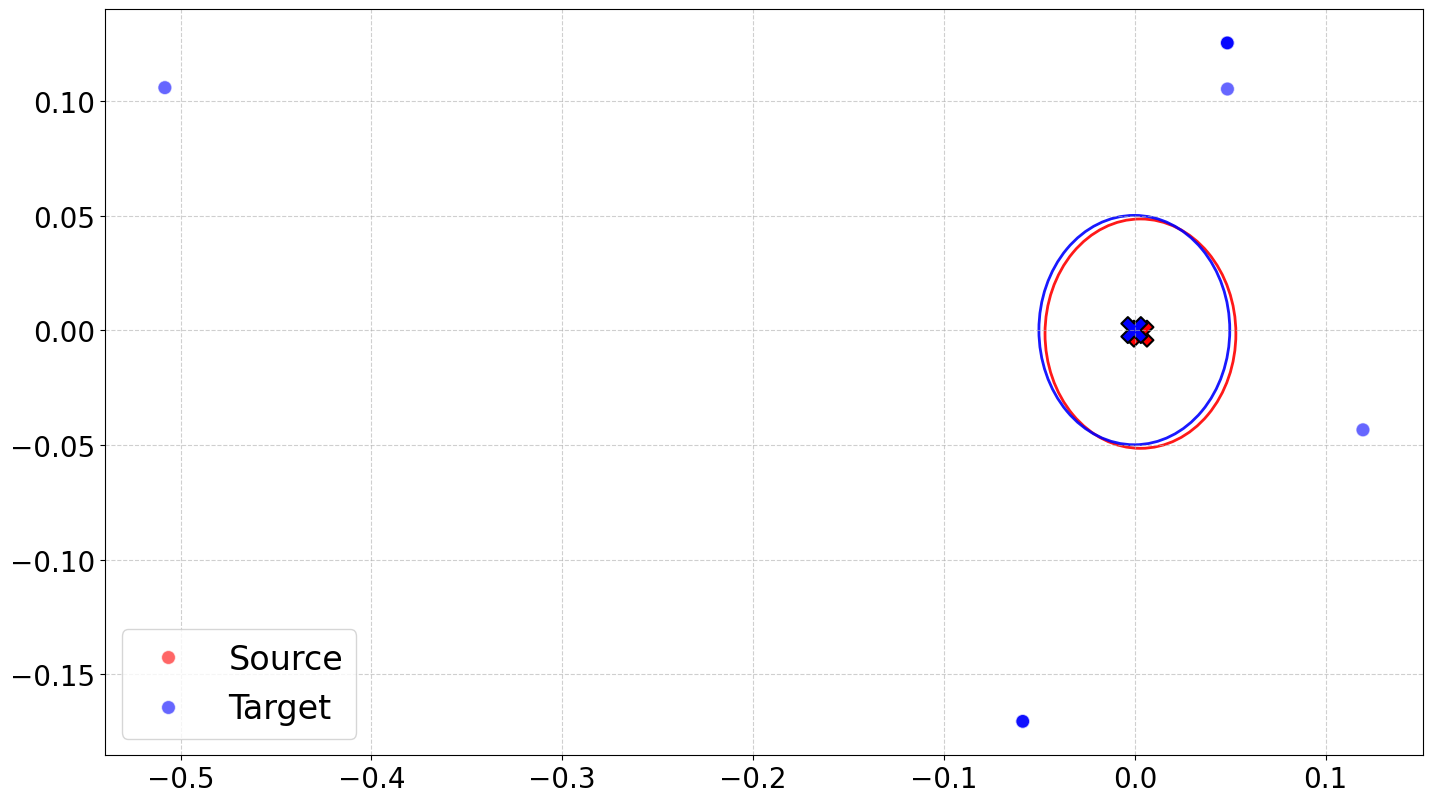}
        \caption{What was the name of the University of Detroit's halfback in the 1929 football season?}
    \end{subfigure}
    \begin{subfigure}[t]{0.32\linewidth}
        \includegraphics[width=\linewidth]{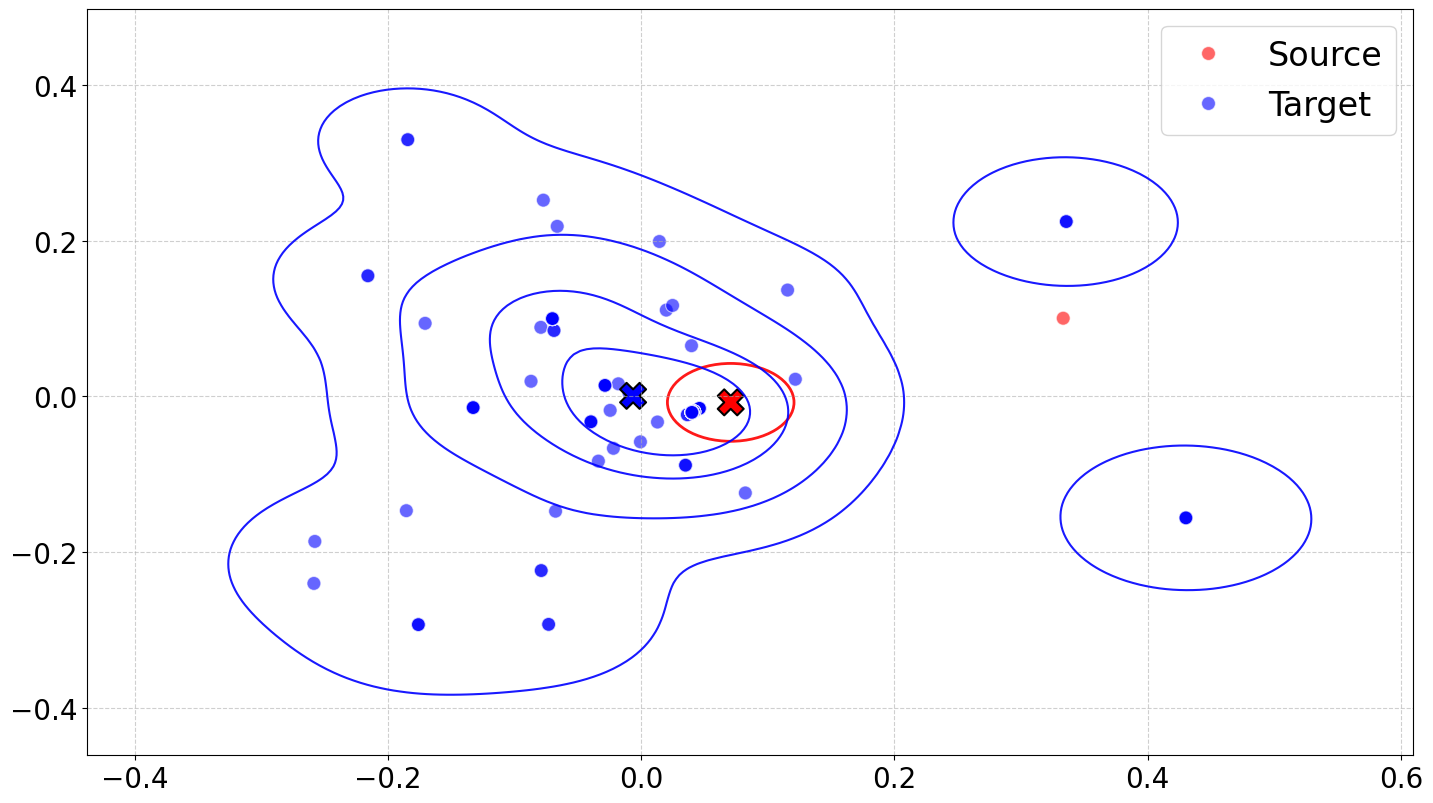}
        \caption{Alongside which council does the Israeli Coastal Environment Preservation Committee operate?}
    \end{subfigure}
    
    \vspace{0.5cm} 

    \begin{subfigure}[t]{0.32\linewidth}
        \includegraphics[width=\linewidth]{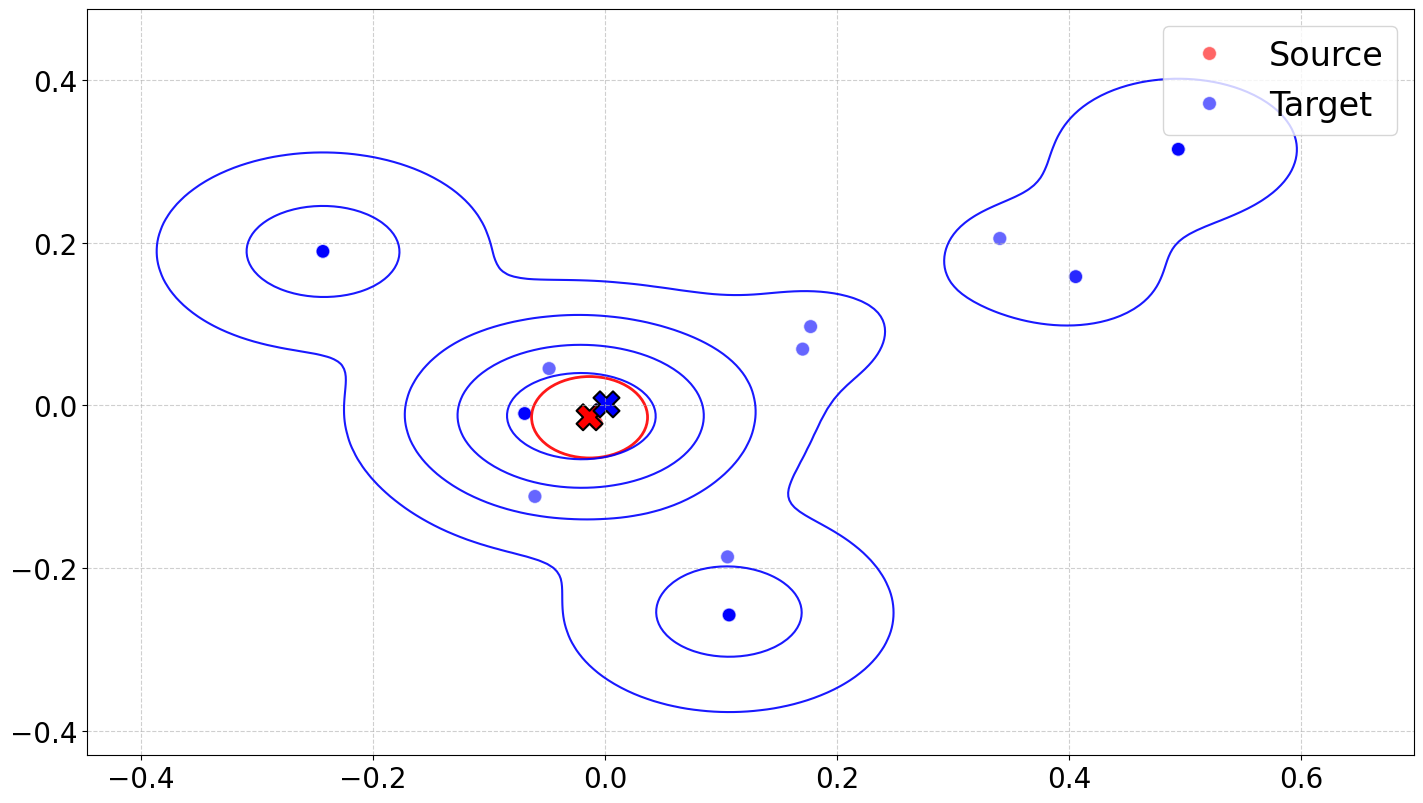} 
        \caption{Who was the first Grand Commander of Eastern Wu?}

    \end{subfigure}
    \begin{subfigure}[t]{0.32\linewidth}
        \includegraphics[width=\linewidth]{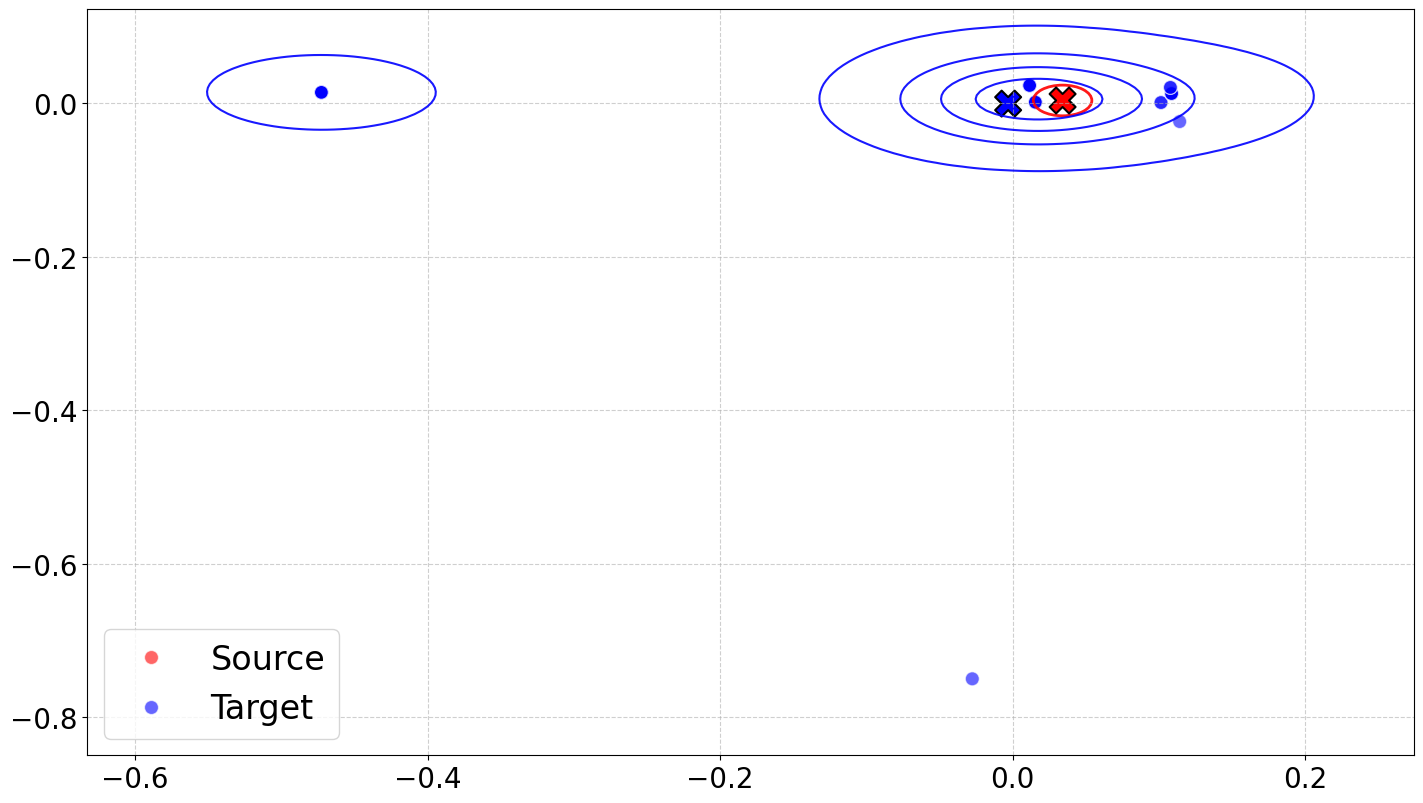} 
        \caption{Who designed the Olivetti DL typewriter?}
    \end{subfigure}
    \begin{subfigure}[t]{0.32\linewidth}
        \includegraphics[width=\linewidth]{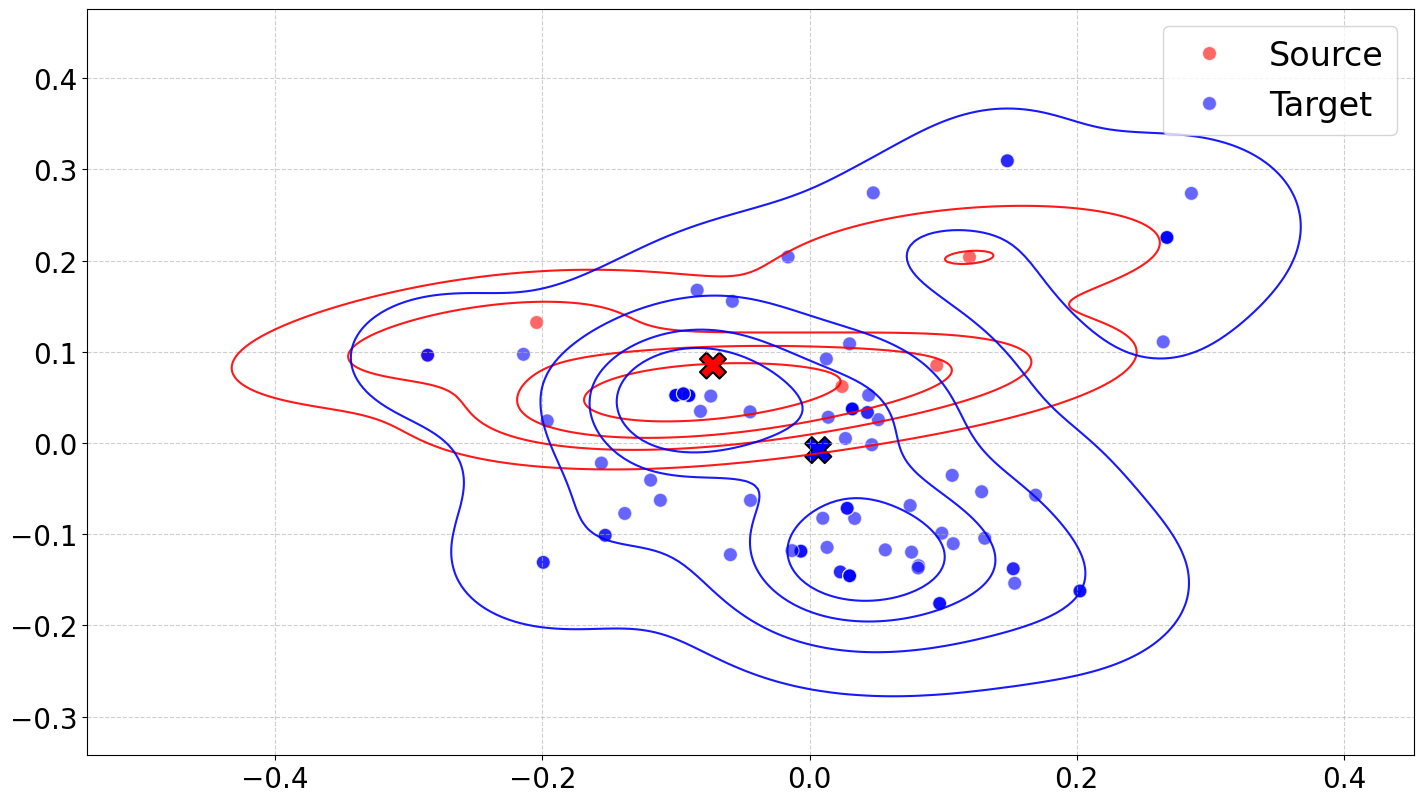} 
        \caption{What colors do the members of the Landsmannschaft Saxonia wear?}
    \end{subfigure}

    \caption{Additional results from Figure~\ref{fig:problem_illustrated}b on \eclektic{}.}
    \label{fig:main_b_extended} 
\end{figure}

\begin{figure}[htb] 
    \centering 

    \begin{subfigure}[t]{0.32\linewidth}
        \includegraphics[width=\linewidth]{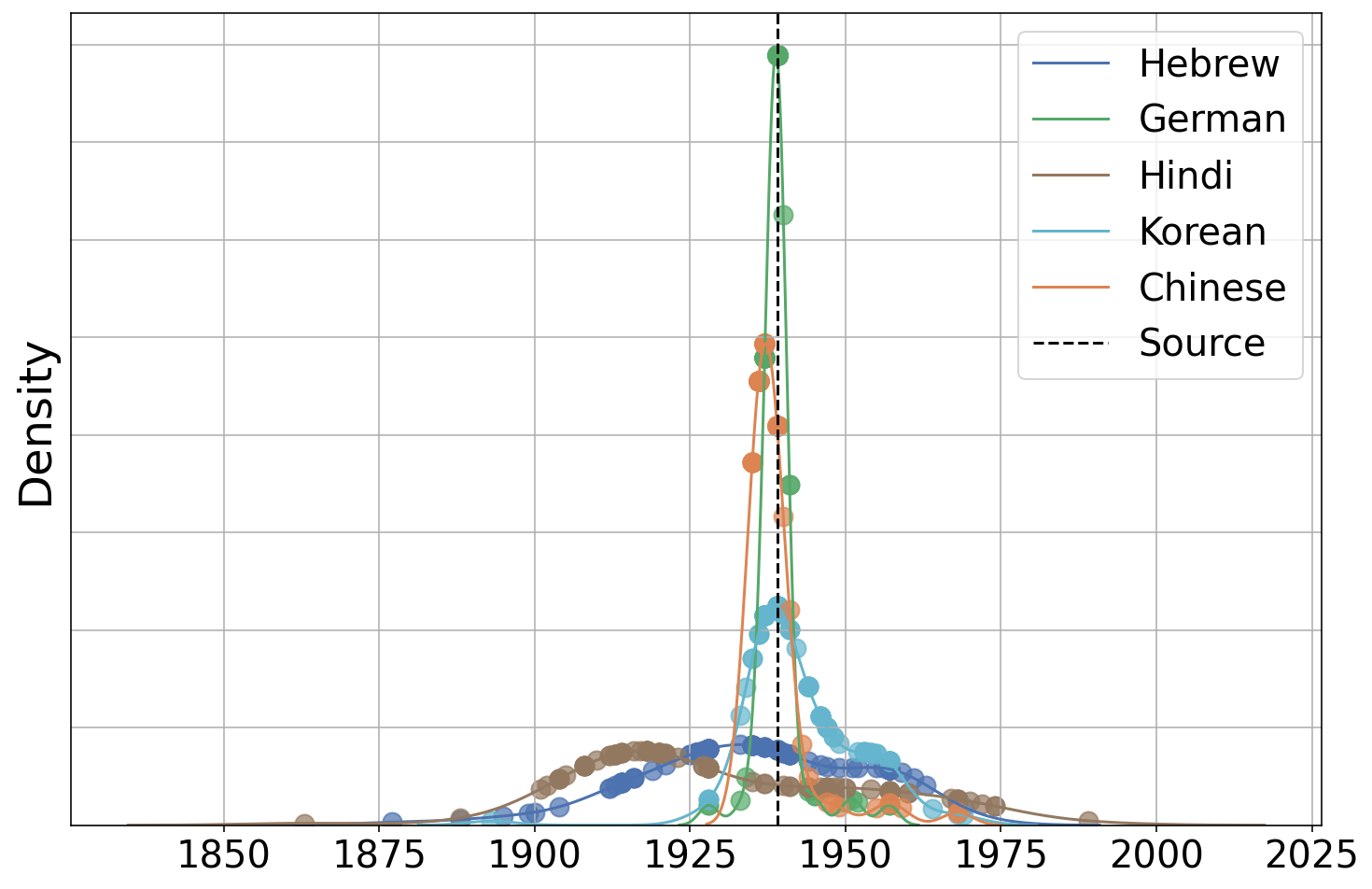}
        \caption{In which year was Siegfried Loyda born?}
    \end{subfigure}
    \begin{subfigure}[t]{0.32\linewidth}
        \includegraphics[width=\linewidth]{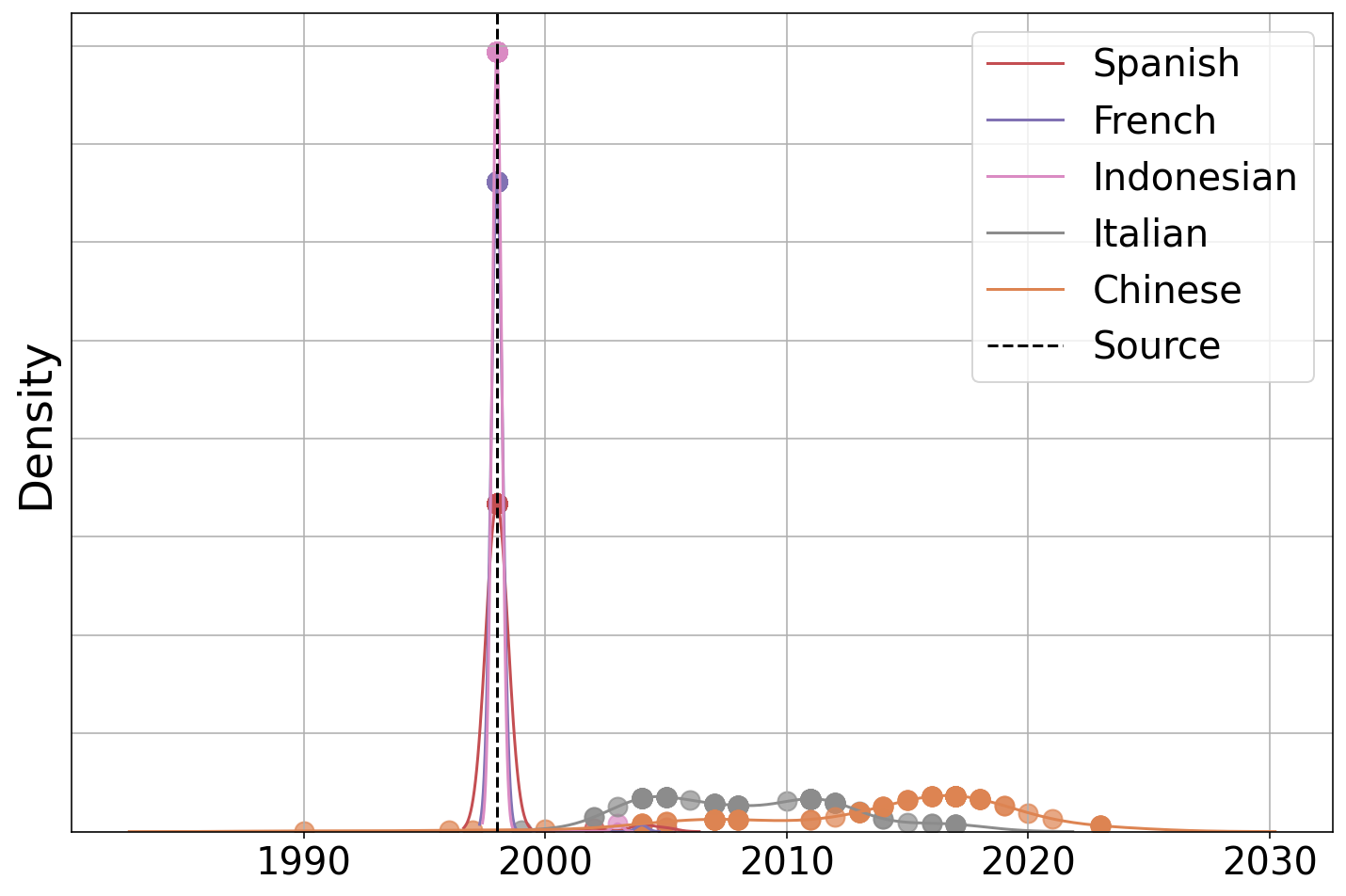}
        \caption{In which year was the first edition of the Rock Basement festival held?}
    \end{subfigure}
    \begin{subfigure}[t]{0.32\linewidth}
        \includegraphics[width=\linewidth]{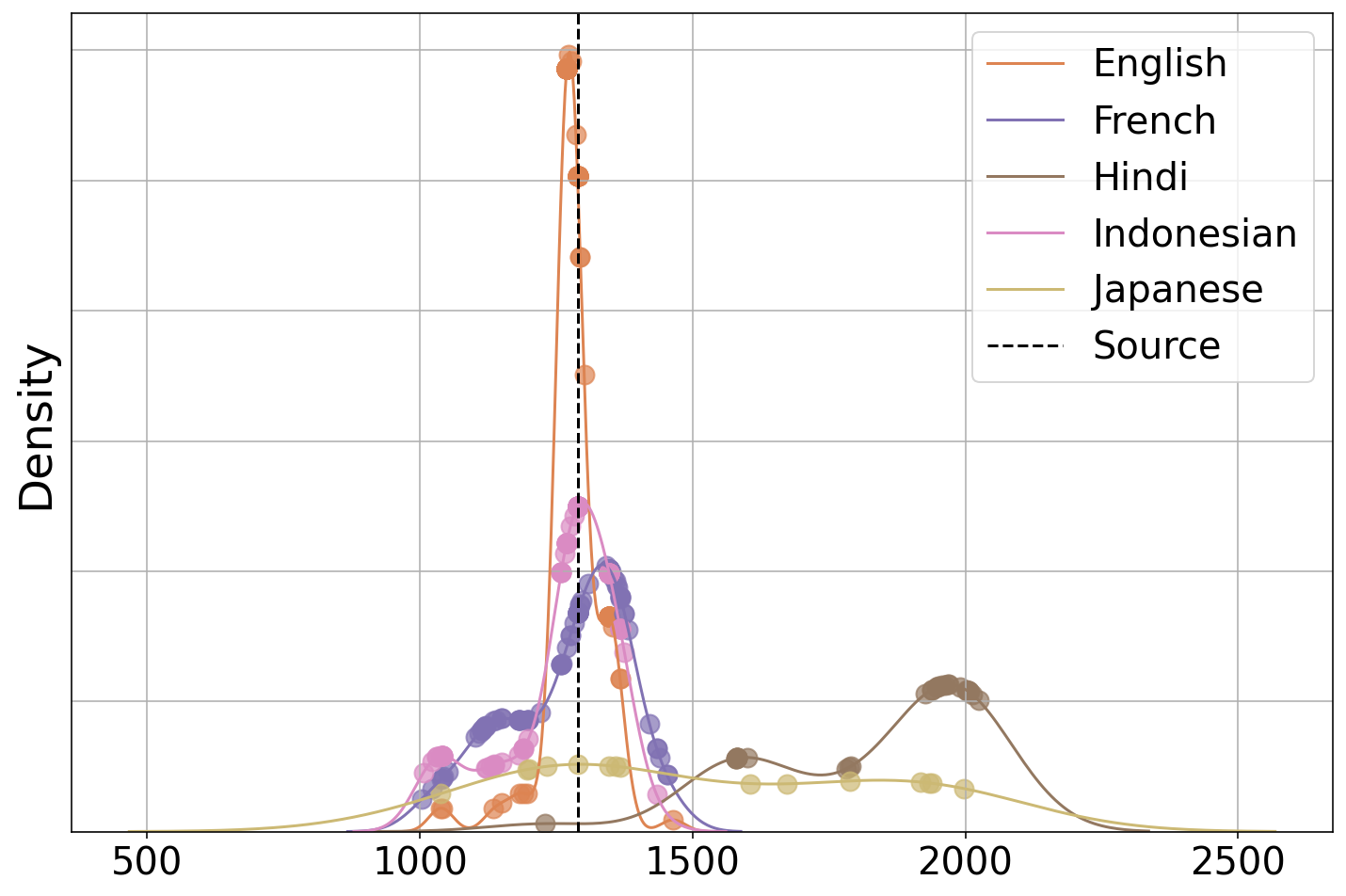}
        \caption{In what year was the Sarwadharma Inscription issued?}
    \end{subfigure}
    
    \vspace{0.5cm} 

    \begin{subfigure}[t]{0.32\linewidth}
        \includegraphics[width=\linewidth]{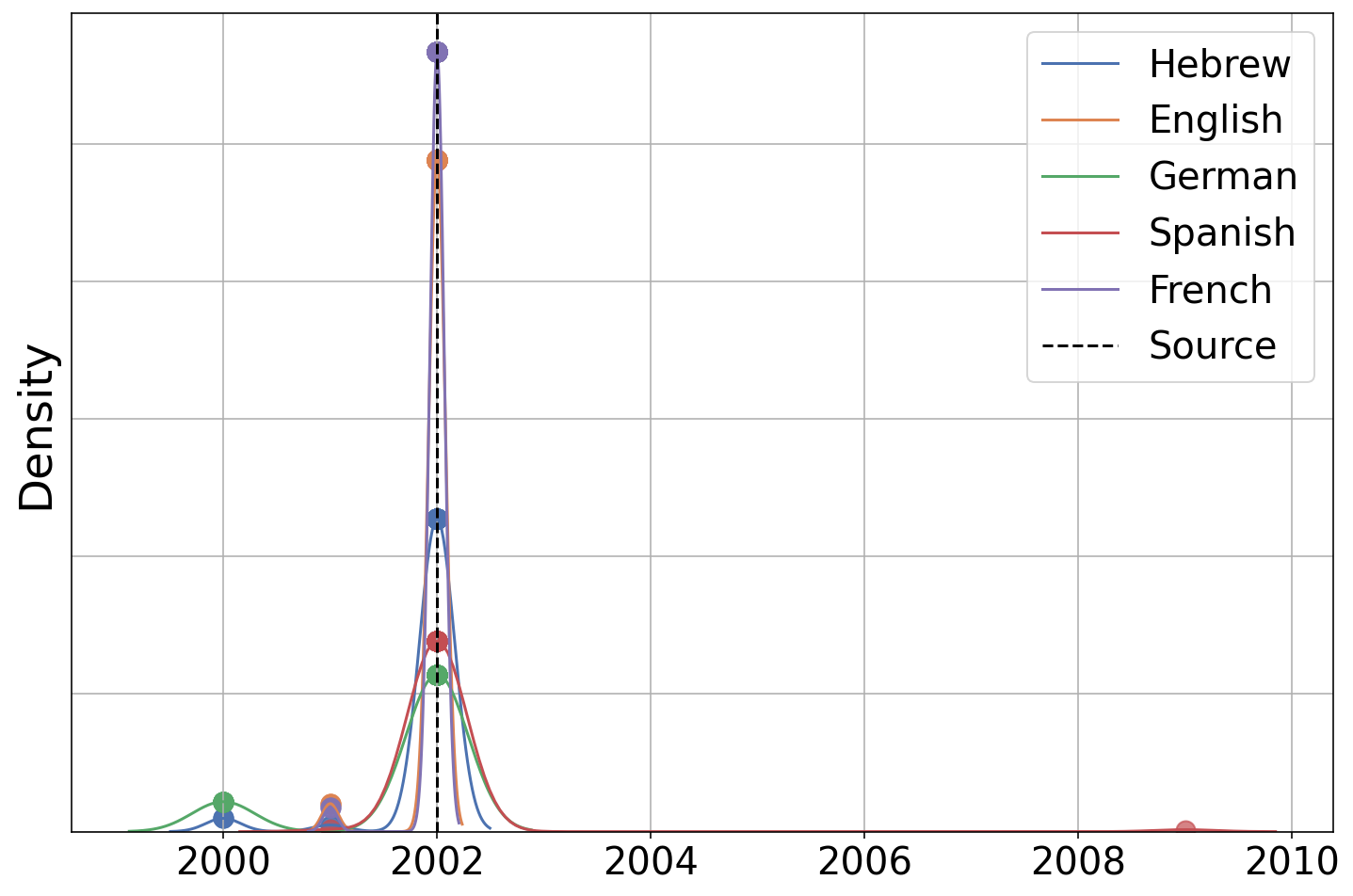} 
        \caption{In which year were the "Girotondi" movements born?}

    \end{subfigure}
    \begin{subfigure}[t]{0.32\linewidth}
        \includegraphics[width=\linewidth]{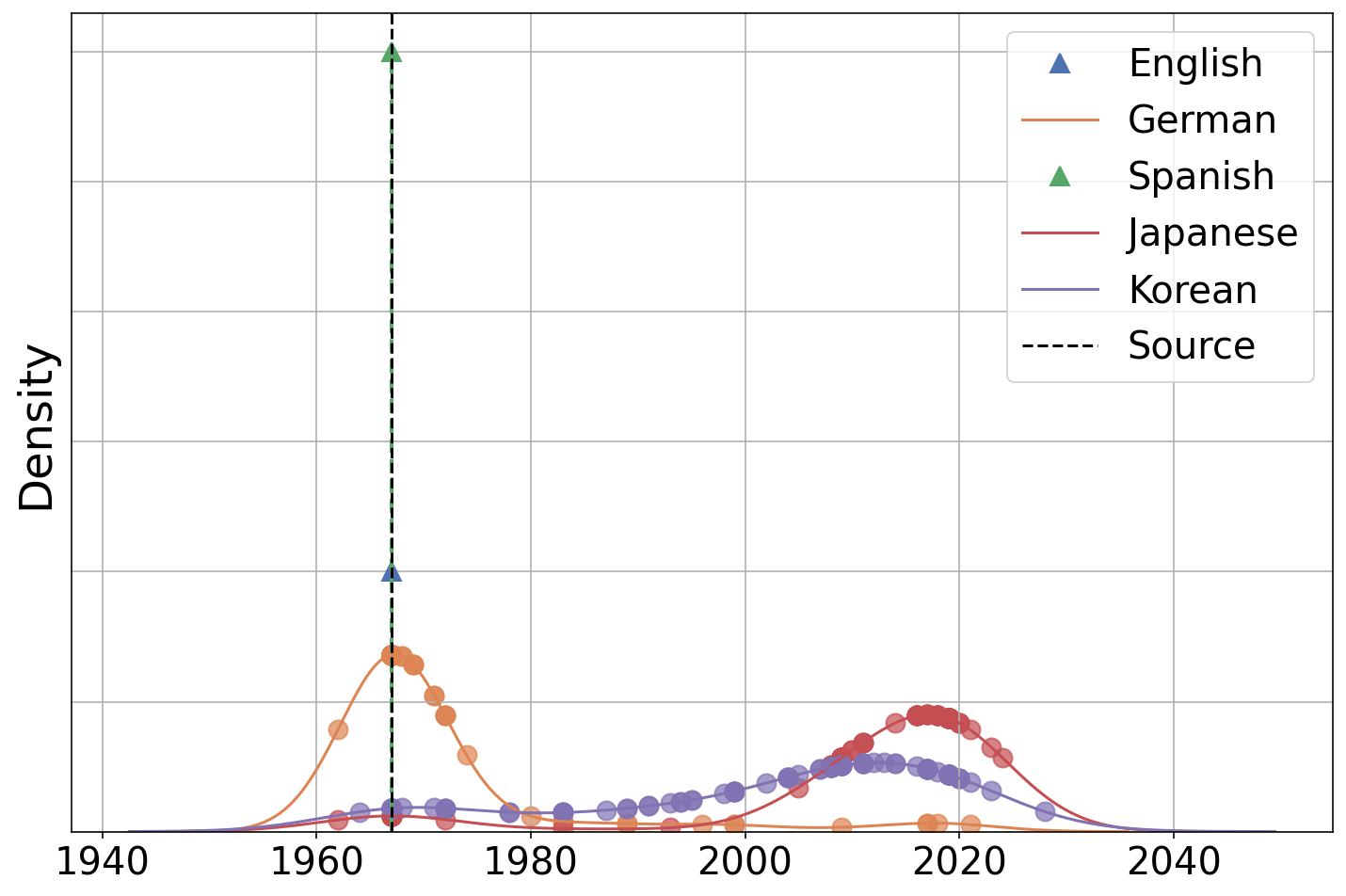} 
        \caption{In what year was the film "Three Days and a Child" released with Judith Soleh?}
    \end{subfigure}
    \begin{subfigure}[t]{0.32\linewidth}
        \includegraphics[width=\linewidth]{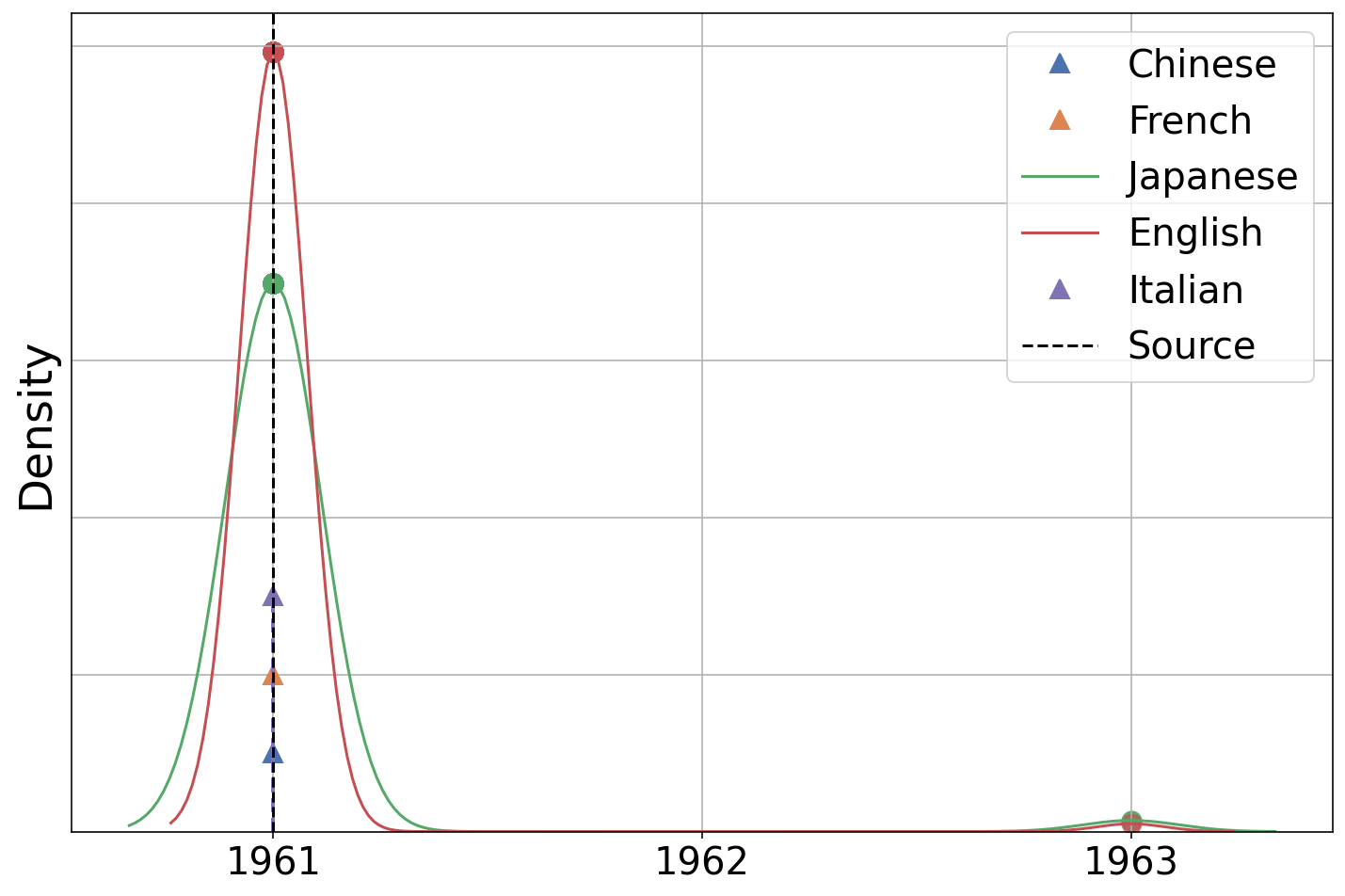} 
        \caption{When was the Indian Institute of Technology, Delhi established?}
    \end{subfigure}

    \caption{Additional results from Figure~\ref{fig:problem_illustrated}c on \eclektic{}.}
    \label{fig:main_c_extended} 
\end{figure}

\section{Do our insights from Section~\ref{sec:ensemble_expts} hold for any languages?}
\label{sec:fine_ensemble_expts}
In this section, we investigate if our analysis and insights hold at a finer level of cross-lingual transfer: high-to-low resource and vice-versa. We picked two languages from \eclektic{}: \texttt{en, zh} as high-resource and two other languages: \texttt{he, id} as low-resource. We followed the statistics for Wikipedia pages mentioned at \textit{\href{https://en.wikipedia.org/wiki/List_of_Wikipedias}{WikiSources}} to ascertain high and low resource languages.

We replicate our main findings: Figure~\ref{fig:point_robust}, Table~\ref{tab:results:input_ensemble} for all combinations of high and low resource languages. All the results are consistent with the analysis in Section~\ref{sec:ensemble_expts} with the only difference on the evaluation subset used for aggregation. For instance, high$\rightarrow$low analysis considers only the examples that originated in a high resource language: \texttt{en} or \texttt{zh} and being queried in a low resource language: \texttt{he} or \texttt{id}. Please refer to Figures~\ref{fig:g25flhilo}, \ref{fig:g25prohilo}, \ref{fig:gpt5minihilo}, \ref{fig:gpt5hilo} and \ref{fig:drhilo}. and Tables~\ref{tab:hilo}, \ref{tab:hihi}, \ref{tab:lohi} and \ref{tab:lolo}.

\begin{figure}[htb]
    \includegraphics[width=\linewidth]{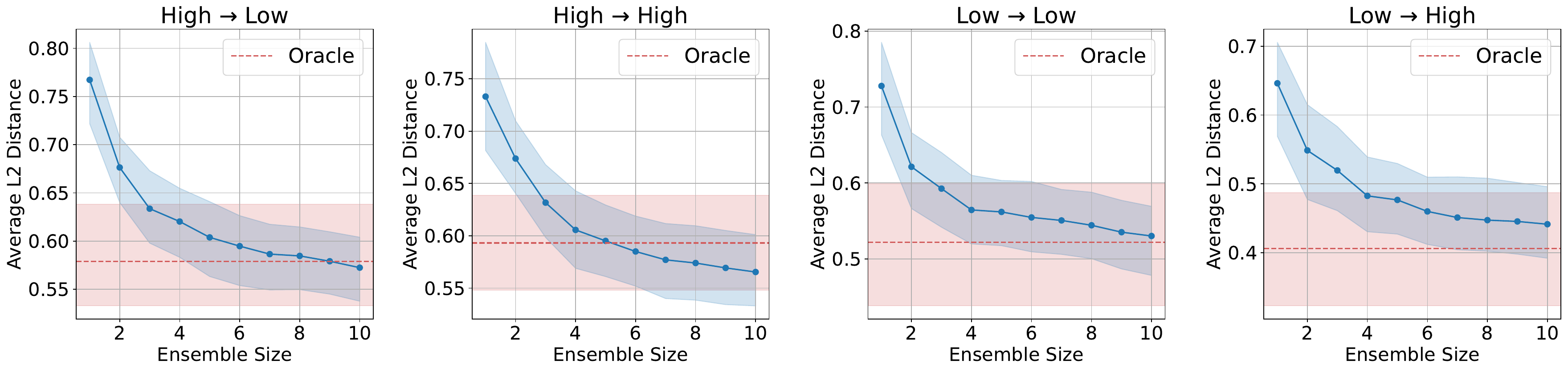}
    \caption{Reproducing results from Figure~\ref{fig:point_robust} on \eclektic{} for Gemini 2.5 Flash.}
    \label{fig:g25flhilo}
\end{figure}
\begin{figure}[htb]
    \includegraphics[width=\linewidth]{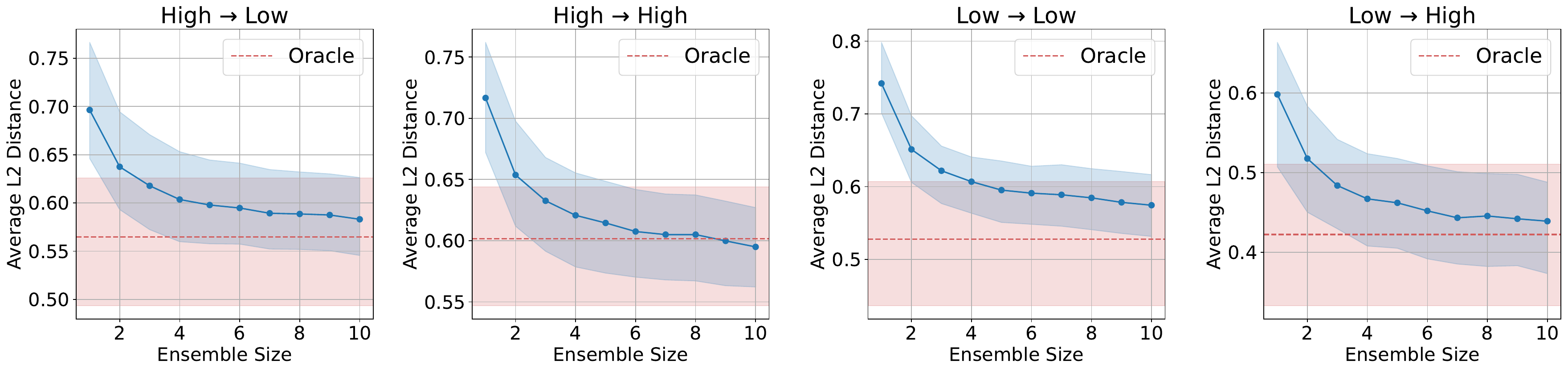}
    \caption{Reproducing results from Figure~\ref{fig:point_robust} on \eclektic{} for Gemini 2.5 Pro.}
    \label{fig:g25prohilo}
\end{figure}
\begin{figure}[htb]
    \includegraphics[width=\linewidth]{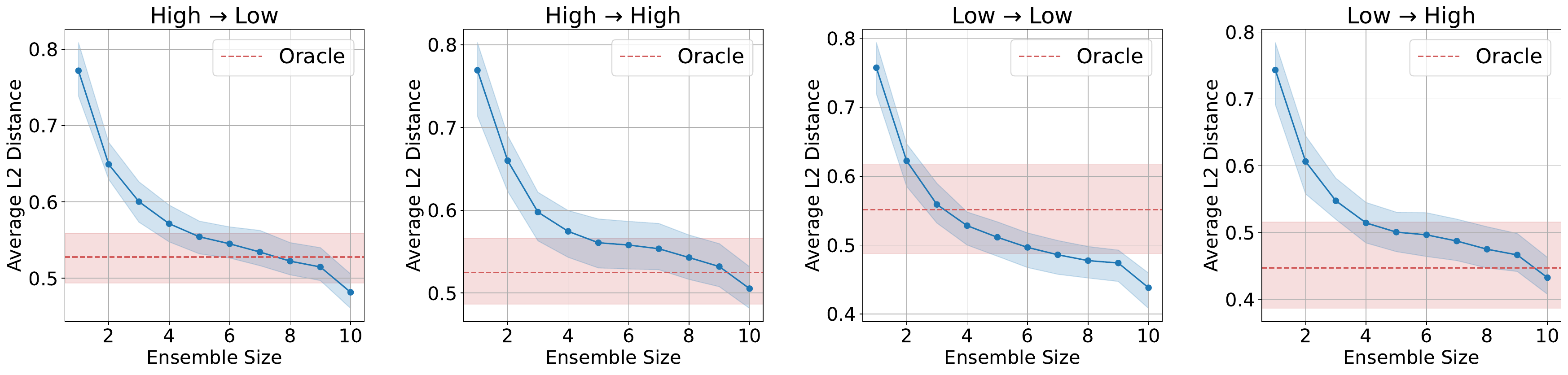}
    \caption{Reproducing results from Figure~\ref{fig:point_robust} on \eclektic{} for GPT-5 mini.}
    \label{fig:gpt5minihilo}
\end{figure}
\begin{figure}[htb]
    \includegraphics[width=\linewidth]{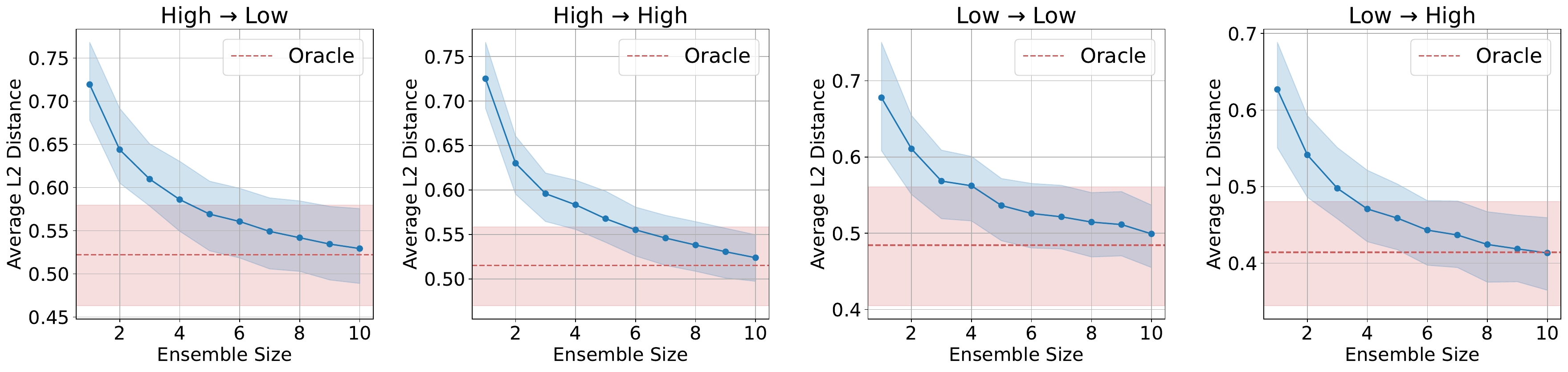}
    \caption{Reproducing results from Figure~\ref{fig:point_robust} on \eclektic{} for GPT-5.}
    \label{fig:gpt5hilo}
\end{figure}
\begin{figure}[htb]
    \includegraphics[width=\linewidth]{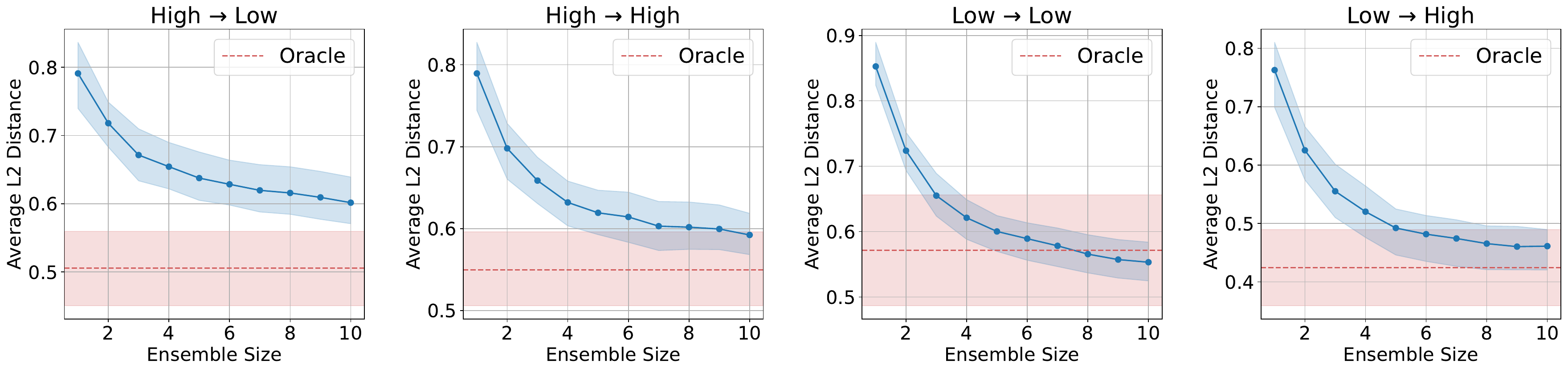}
    \caption{Reproducing results from Figure~\ref{fig:point_robust} on \eclektic{} for DeepSeek.}
    \label{fig:drhilo}
\end{figure}

\begin{table}[htb]
    \centering
    \begin{tabular}{l|c|c|c|c|c|c}
          & G-2.5-Flash & G-2.5-Pro & GPT-5-mini & GPT-5 & Deepseek & Gem-3-27B \\ \hline
 Baseline & 38.7        & 43.7      & 28.6       & 45.3  & 23.4     & 13.0      \\ \hline
 TrEn-1   & 44.2        & 51.9      & 31.2       & 48.3  & 34.1     & 15.6      \\
 TrEn-3   & 42.0        & 52.4      & 32.5       & 49.7  & 27.3     & 14.3      \\
 TrEn-5   & 48.1        & 51.5      & 35.1       & 50.9  & 30.7     & 15.6      \\ 
\end{tabular}
    \caption{Source-Target transfer scores for \eclektic{} for High resource language to Low resource language.}
    \label{tab:hilo}
\end{table}
\begin{table}[htb]
    \centering
    \begin{tabular}{l|c|c|c|c|c|c}
          & G-2.5-Flash & G-2.5-Pro & GPT-5-mini & GPT-5 & Deepseek & Gem-3-27B \\ \hline
 Baseline & 39.9        & 46.8      & 28.6       & 45.8  & 27.3     & 14.3      \\ \hline
 TrEn-1   & 42.9        & 51.9      & 29.9       & 48.0  & 25.6     & 14.3      \\
 TrEn-3   & 42.9        & 50.6      & 32.5       & 50.9  & 32.6     & 15.6      \\
 TrEn-5   & 51.9        & 54.5      & 36.4       & 50.9  & 27.9     & 18.2      \\ 
\end{tabular}
    \caption{Source-Target transfer scores for \eclektic{} for High resource language to High resource language.}
    \label{tab:hihi}
\end{table}
\begin{table}[htb]
    \centering
    \begin{tabular}{l|c|c|c|c|c|c}
          & G-2.5-Flash & G-2.5-Pro & GPT-5-mini & GPT-5 & Deepseek & Gem-3-27B \\ \hline
 Baseline & 28.7        & 36.6      & 20.2       & 38.5  & 13.1     & 8.5       \\ \hline
 TrEn-1   & 27.7        & 33.3      & 18.8       & 28.8  & 13.4     & 10.3      \\
 TrEn-3   & 29.1        & 36.6      & 22.1       & 30.6  & 13.7     & 11.3      \\
 TrEn-5   & 34.7        & 39.4      & 18.3       & 36.6  & 13.7     & 12.7      \\ 
\end{tabular}
    \caption{Source-Target transfer scores for \eclektic{} for Low resource language to High resource language.}
    \label{tab:lohi}
\end{table}
\begin{table}[htb]
    \centering
    \begin{tabular}{l|c|c|c|c|c|c}
          & G-2.5-Flash & G-2.5-Pro & GPT-5-mini & GPT-5 & Deepseek & Gem-3-27B \\ \hline
 Baseline & 29.1        & 37.8      & 26.8       & 46.6  & 8.5      & 8.5       \\ \hline
 TrEn-1   & 20.7        & 23.2      & 14.6       & 26.4  & 10.2     & 8.5       \\
 TrEn-3   & 25.6        & 28.0      & 20.7       & 25.0  & 8.3      & 8.5       \\
 TrEn-5   & 26.8        & 36.6      & 14.6       & 31.0  & 12.5     & 11.0      \\ 
\end{tabular}
    \caption{Source-Target transfer scores for \eclektic{} for Low resource language to Low resource language.}
    \label{tab:lolo}
\end{table}

\section{Do our insights from Section~\ref{sec:confidence_vs_gaps} hold at a finer level of cross-lingual transfer?}
\label{sec:fine_confidence_vs_gaps}
In this section we further experiment and solidify our findings from Section~\ref{sec:confidence_vs_gaps}. We plot trend of source-target agreement with source confidence for 4 combinations with source and target language either being high resource or low resource. Here we use the full-set of \eclektic{} dataset and identified English, German, French, Spanish, Chinese and Italian as high resource languages and Portuguese, Japanese, Korean, Hebrew, Hindi and Indonesian as low resource languages. Our findings confirm that the trend still holds for different sets of language pairs. Please refer Figure~\ref{fig:src_tgt_langpairs}.
\begin{figure}[htb] 
    \centering 

    \begin{subfigure}[t]{0.45\linewidth}
        \includegraphics[width=\linewidth]{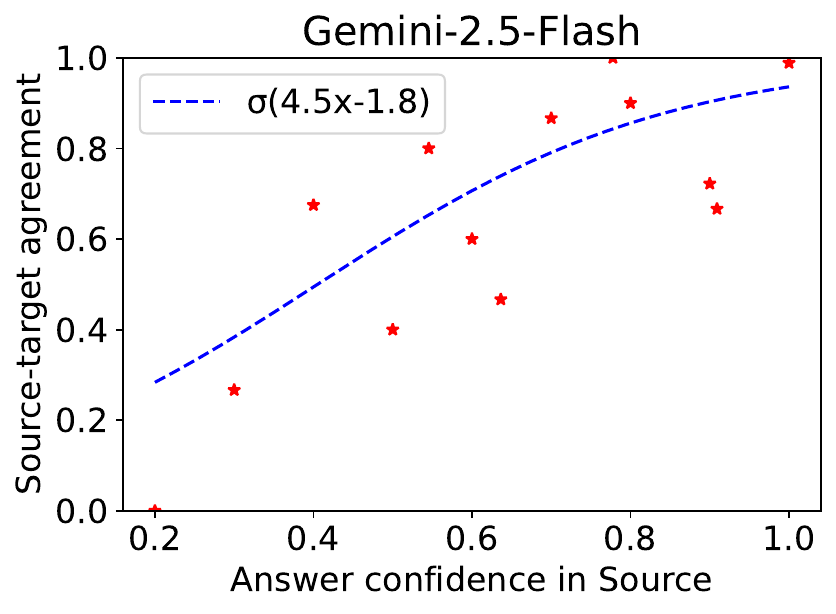}
        \caption{High Resource Source Language, High Resource Target Language}
    \end{subfigure}
    \hfill 
    \begin{subfigure}[t]{0.45\linewidth}
        \includegraphics[width=\linewidth]{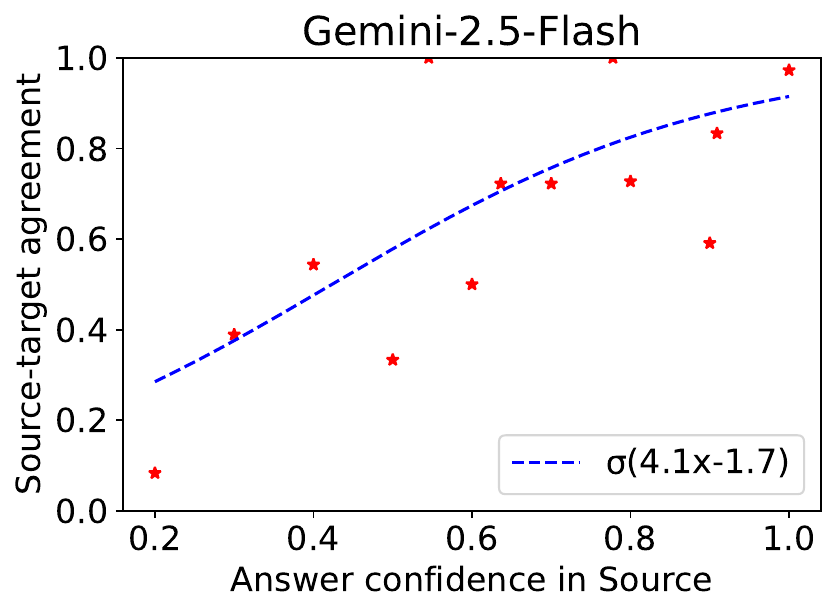}
        \caption{High Resource Source Language, Low Resource Target Language}
    \end{subfigure}

    \vspace{0.5cm} 

    \begin{subfigure}[t]{0.45\linewidth}
        \includegraphics[width=\linewidth]{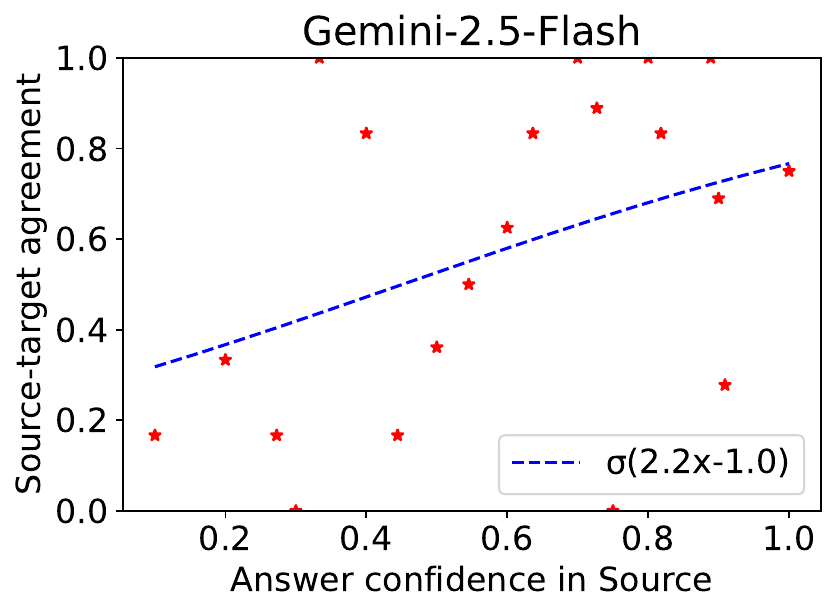}
        \caption{Low Resource Source Language, High Resource Target Language}
    \end{subfigure}
    \hfill 
    \begin{subfigure}[t]{0.45\linewidth}
        \includegraphics[width=\linewidth]{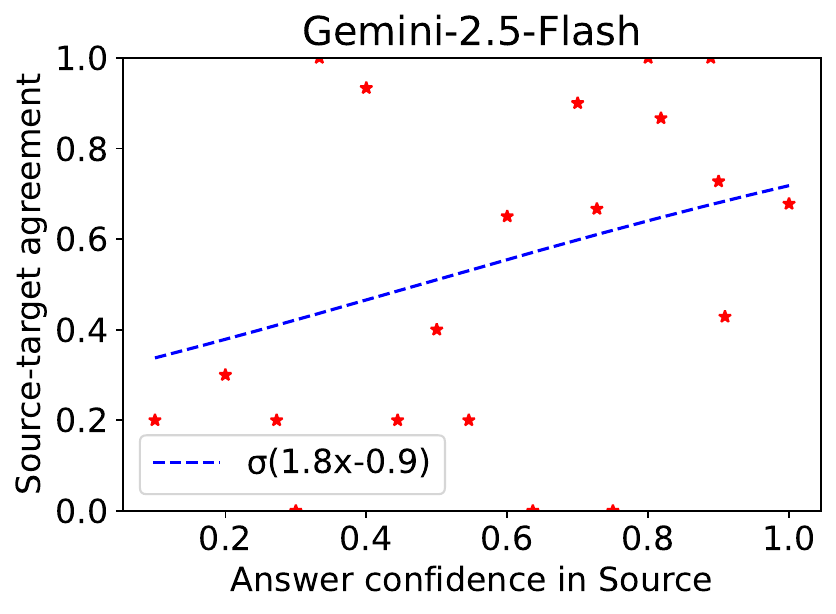}
        \caption{Low Resource Source Language, Low Resource Target Language}
    \end{subfigure}

    \caption{Additional results from Section~\ref{sec:confidence_vs_gaps} on \neclektic{}.} 
    \label{fig:src_tgt_langpairs}
\end{figure}

\section{Miscellaneous}
\label{sec:misc}
\subsection{Response matching prompt}
\label{sec:mode_match_prompt}
Please find the full prompt used for matching strings potentially different languages in \texttt{response\_matching\_prompt.txt} in supplementary material.  

\subsection{Response Summarizer prompt}
\label{sec:response_summarizer_prompt}
Please find the full prompt used for summarizing a list of strings into their unique values and counts in \texttt{response\_summarizer\_prompt.txt} in supplementary material.

\subsection{Autochecker prompt}
\label{sec:autochecker}
Please find the full prompt used for autochecking if a response matched the reference in \texttt{autochecker\_prompt.txt} attached in supplementary material.
The provided autochecker prompt is for the response: {\it A: L'ordre de Santiago.} and reference: {\it ordre de santiago}.


\subsection{pi estimation for \mmlu{}}
\label{sec:misc:mmlu_soft_approx}
We provide additional details of estimation for $\pi$ with \mmlu{} described in Section~\ref{sec:ensemble_expts}. 
We estimate $\pi$ for \mmlu{} as one minus fraction of examples with best answer mismatch between source and target. We denote the probability distribution for an $n^{th}$ example in source and target with $p^{(n)}, q^{(n)}$ respectively. We use a soft score for mismatch as described next. 
\begin{align*}
    &i = \arg\max_k p^{(n)}_k, \quad j = \arg\max_k q^{(n)}_k,\\
    &\text{mismatch}^{(n)} = \begin{cases}
        0 & \text{if } i = j, \\
        |p^{(n)}_{i} - q^{(n)}_{i}| & \text{otherwise},
    \end{cases}\\
    &\pi = 1 - \frac{1}{N}\sum_n \text{mismatch}^{(n)}.
\end{align*}

\subsection{Analysis of examples with bias}
\label{sec:misc:translation_errs}
In this section, we spot-check a few examples that are biased even after ensembling, and found them to contain translation errors.  

In \eclektic{} dataset we observed translations errors such as in Figure~\ref{fig:main_c_extended}(b) when the Chinese question was back translated into English, we observed that "Rock Basement" was translated as "Rock Cellar" which led to erroneous answers, confirmed upon passing "Rock Basement" in the question. Similarly in Figure~\ref{fig:main_c_extended}(c) upon translating Japanese question back to English we find that the word "issued" is replaced by "published" leading to erroneous answers.

Similarly, in the \mmlu{} dataset, we observe for the statement-based questions, where the task is to identify if a set of statements are "Right" or "Wrong", often their translations in target languages when back translated to english can have varied meanings like "bad", which take away from the semantic meaning of the answer. On performing a spot-check we observed multiple instances of such statement-based questions.

\subsection{Extended Results}
Below we present individual source and target accuracies in contrast to one conflated measurement reported in Table~\ref{tab:results:input_ensemble}.

       

\begin{table}[htb]
    \centering
    \begin{tabular}{l|c|c|c|c|c}
    \toprule
    & G-2.5-Flash & G-2.5-Pro & GPT-5-mini & GPT-5 & Deepseek \\ 
    \midrule
         Baseline & 59.9, 49.0  & 67.7, 58.9 & 40.1, 36.3 & 64.6, 58.6 & 44.3, 30.9 \\ 
    \midrule
         TEA & 58.9, 49.6 & 70.3, 61.2 & 37.8, 35.2 & 61.7, 58.6 & 52.6, 42.4 \\ 
    \midrule
         TrEn-1 & 58.9, 51.6 & 63.0, 63.0 & 40.4, 39.9 & 65.6, 60.9 & 48.2, 38.4 \\ 
         TrEn-3 & 58.3, 53.6 & 62.8, 62.1 & 39.8, 41.5 & 64.6, 62.5 & 51.0, 42.9 \\ 
         \rowcolor{LightLimeGreen}
         TrEn-5 & 57.6, 54.5 & 66.9, 63.7 & 39.8, 41.5 & 65.9, 62.5 &  51.4, 42.9\\ 
    \bottomrule
    \end{tabular}
    \caption{We show Source-Target accuracies for best ablations from Table~\ref{tab:results:input_ensemble}. The format is Src acc, Tgt acc. The standard deviation is approximately 2.5 for Src and 0.8 for Tgt across all measurements.}
    \label{tab:results:inp_ens_acc}
\end{table}

\section{Computational Cost}
\label{appendix:computational_cost}
All the models that we used in this paper are available publicly via API calls and approximately costed 50k USD. We report results for all the experiments that were conducted and do not discard any results.

\section{Broader Societal Impacts}
\label{appendix:societal}
Our work focuses on analysis and mitigation strategies for cross-lingual gaps in modern LLMs. This is a step towards making LLMs more equitable and catering to people around the world by providing the same experience irrespective of language.
\clearpage
\section*{NeurIPS Paper Checklist}

\begin{enumerate}

\item {\bf Claims}
    \item[] Question: Do the main claims made in the abstract and introduction accurately reflect the paper's contributions and scope?
    \item[] Answer: \answerYes{} 
    \item[] Justification: We formalize the theoretical framework in Section~\ref{sec:model} and Appendix~\ref{sec:proofs}. We also verify our theoretical framework empirically in Section~\ref{sec:results} to show it is grounded in real models.

\item {\bf Limitations}
    \item[] Question: Does the paper discuss the limitations of the work performed by the authors?
    \item[] Answer: \answerYes{} 
    \item[] Justification: Yes, we discuss the limitations in Section~\ref{sec:conclusion}.

\item {\bf Theory assumptions and proofs}
    \item[] Question: For each theoretical result, does the paper provide the full set of assumptions and a complete (and correct) proof?
    \item[] Answer: \answerYes{}{} 
    \item[] Justification: We provide the full set of assumptions and complete proofs in Section~\ref{sec:model} and Appendix~\ref{sec:proofs}.

    \item {\bf Experimental result reproducibility}
    \item[] Question: Does the paper fully disclose all the information needed to reproduce the main experimental results of the paper to the extent that it affects the main claims and/or conclusions of the paper (regardless of whether the code and data are provided or not)?
    \item[] Answer: \answerYes{} 
    \item[] Justification: We provide full details of the datasets, models, hyperparameters and evaluation setup in Section~\ref{sec:expt_setup}.

\item {\bf Open access to data and code}
    \item[] Question: Does the paper provide open access to the data and code, with sufficient instructions to faithfully reproduce the main experimental results, as described in supplemental material?
    \item[] Answer: \answerNo{} 
    \item[] Justification: All the datasets and models are publicly available. We provide detailed experimental setup (prompts, hyperparameters etc.) in Section ~\ref{sec:expt_setup} for anyone to reproduce.

\item {\bf Experimental setting/details}
    \item[] Question: Does the paper specify all the training and test details (e.g., data splits, hyperparameters, how they were chosen, type of optimizer) necessary to understand the results?
    \item[] Answer: \answerYes{} 
    \item[] Justification: We specify the models, datasets, languages, and evaluation metrics in Section~\ref{sec:expt_setup} and Appendix~\ref{sec:dataset_details}, along with hyperparameters (like temperature and ensemble size) in Section~\ref{sec:ensemble_expts}.

\item {\bf Experiment statistical significance}
    \item[] Question: Does the paper report error bars suitably and correctly defined or other appropriate information about the statistical significance of the experiments?
    \item[] Answer: \answerYes{} 
    \item[] Justification: We include confidence intervals in our main plots (e.g., Figures~\ref{fig:point_robust},~\ref{fig:neclektic_acc}) and report standard deviations where appropriate.

\item {\bf Experiments compute resources}
    \item[] Question: For each experiment, does the paper provide sufficient information on the computer resources (type of compute workers, memory, time of execution) needed to reproduce the experiments?
    \item[] Answer: \answerYes{} 
    \item[] Justification: The models that we used in this paper are available publicly via API calls we mention related costs in Appendix~\ref{appendix:computational_cost}.
    
\item {\bf Code of ethics}
    \item[] Question: Does the research conducted in the paper conform, in every respect, with the NeurIPS Code of Ethics \url{https://neurips.cc/public/EthicsGuidelines}?
    \item[] Answer: \answerYes{} 
    \item[] Justification: Our work follows the NeurIPS Code of Ethics.

\item {\bf Broader impacts}
    \item[] Question: Does the paper discuss both potential positive societal impacts and negative societal impacts of the work performed?
    \item[] Answer: \answerYes{} 
    \item[] Justification: Our paper investigates cross-lingual gaps with the broader goal of equitable access to LLMs. Please refer Appendix~\ref{appendix:societal}.    
    
\item {\bf Safeguards}
    \item[] Question: Does the paper describe safeguards that have been put in place for responsible release of data or models that have a high risk for misuse (e.g., pre-trained language models, image generators, or scraped datasets)?
    \item[] Answer: \answerNA{} 
    \item[] Justification: We do not train or release any new models or datasets that carry a high risk of misuse.

\item {\bf Licenses for existing assets}
    \item[] Question: Are the creators or original owners of assets (e.g., code, data, models), used in the paper, properly credited and are the license and terms of use explicitly mentioned and properly respected?
    \item[] Answer: \answerNA{}{} 
    \item[] Justification: We cite all the datasets (ECLeKTic, MMLU, MultiLoKo) and models used in our evaluations. But we have not explicitly mentioned the licenses for each of the assets.   

\item {\bf New assets}
    \item[] Question: Are new assets introduced in the paper well documented and is the documentation provided alongside the assets?
    \item[] Answer: \answerNA{}  
    \item[] Justification: The paper does not introduce new datasets or models.

\item {\bf Crowdsourcing and research with human subjects}
    \item[] Question: For crowdsourcing experiments and research with human subjects, does the paper include the full text of instructions given to participants and screenshots, if applicable, as well as details about compensation (if any)? 
    \item[] Answer: \answerNA{} 
    \item[] Justification: This work does not involve crowdsourcing or human subjects.

\item {\bf Institutional review board (IRB) approvals or equivalent for research with human subjects}
    \item[] Question: Does the paper describe potential risks incurred by study participants, whether such risks were disclosed to the subjects, and whether Institutional Review Board (IRB) approvals (or an equivalent approval/review based on the requirements of your country or institution) were obtained?
    \item[] Answer: \answerNA{} 
    \item[] Justification: This work does not involve human subjects.

\item {\bf Declaration of LLM usage}
    \item[] Question: Does the paper describe the usage of LLMs if it is an important, original, or non-standard component of the core methods in this research? Note that if the LLM is used only for writing, editing, or formatting purposes and does \emph{not} impact the core methodology, scientific rigor, or originality of the research, declaration is not required.
    \item[] Answer: \answerNA{} 
    \item[] Justification: We do not use LLMs to develop any methods in our research.

\end{enumerate}
\end{document}